\documentclass[11pt]{amsart}
\def\isdraft{0}

\usepackage{mathtools}
\usepackage{fontspec}
\usepackage{unicode-math} % nur mit XeLaTeX/LuaLaTeX
\usepackage{amsmath,amsfonts,amsthm}
\usepackage{amsaddr} % remove in amsbook, JAIR, IOS-Press
\usepackage[dvipsnames]{xcolor}
\usepackage{bbm}
\usepackage{graphicx}
\usepackage{prettyref} % achtung: position dieses kommandos scheint essentiell zu sein
\usepackage[utf8]{inputenc}
\usepackage{enumitem}
\usepackage[hyphens]{url} % remove in JAIR
\usepackage{tikz-cd}
\usepackage{tikz}
\usetikzlibrary{positioning}
\usetikzlibrary{arrows}
\usepackage{bussproofs}
\usepackage[mathscr]{euscript}
\usepackage{hyperref}
\usepackage{a4wide}
\usepackage[color=black,textcolor=white\if\isdraft0,disable\fi]{todonotes}
\usepackage{stackengine}
\usepackage{stmaryrd}
\usepackage{wrapfig}
\usepackage{marginnote}
\usepackage{float}
\graphicspath{{Figures/}}
\usetikzlibrary{arrows,automata,topaths,matrix,positioning,fit}
\tikzset{every state/.style={minimum size=0pt}}
\usepackage{footnote}
\usepackage{tabularx}
\usepackage[normalem]{ulem}
\usepackage{pifont}
\usepackage[most]{tcolorbox}
\usepackage{shuffle}

\newtheorem{theorem}{Theorem}

\newtheorem{corollary}[theorem]{Corollary}
\newtheorem{fact}[theorem]{Fact}
\newtheorem{lemma}[theorem]{Lemma}
\newtheorem{proposition}[theorem]{Proposition}

\theoremstyle{definition} % TPLP remove

\newtheorem{definition}[theorem]{Definition}

\newtheorem{example}[theorem]{Example}

\newtheorem{remark}[theorem]{Remark}

\newrefformat{§}{§\ref{#1}}
\newrefformat{a}{Answer \ref{#1}}
\newrefformat{ana}{Analogy \ref{#1}}
\newrefformat{alg}{Algorithm \ref{#1}}
\newrefformat{ax}{Appendix \ref{#1}}
\newrefformat{cl}{Claim \ref{#1}}
\newrefformat{con}{Conclusion \ref{#1}}
\newrefformat{conv}{Convention \ref{#1}}
\newrefformat{cj}{Conjecture \ref{#1}}
\newrefformat{c}{Corollary \ref{#1}}
\newrefformat{q}{(\ref{#1})}
\newrefformat{e}{Example \ref{#1}}
\newrefformat{exe}{Exercise \ref{#1}}
\newrefformat{d}{Definition \ref{#1}}
\newrefformat{Done}{Done \ref{#1}}
\newrefformat{f}{Fact \ref{#1}}
\newrefformat{fig}{Figure \ref{#1}}
\newrefformat{h}{Hypothesis \ref{#1}}
\newrefformat{idea}{Idea \ref{#1}}
\newrefformat{i}{Item \ref{#1}}
\newrefformat{l}{Lemma \ref{#1}}
\newrefformat{note}{Note \ref{#1}}
\newrefformat{n}{Notation \ref{#1}}
\newrefformat{o}{Observation \ref{#1}}
\newrefformat{pr}{Problem \ref{#1}}
\newrefformat{p}{Proposition \ref{#1}}
\newrefformat{pc}{Pseudocode \ref{#1}}
\newrefformat{Q}{Q. \ref{#1}}
\newrefformat{r}{Remark \ref{#1}}
\newrefformat{ta}{Table \ref{#1}}
\newrefformat{t}{Theorem \ref{#1}}
\newrefformat{T}{Todo \ref{#1}}
\newrefformat{w}{Warning \ref{#1}}

\newcommand{\boxblacktriangle}{\mathrel{\ooalign{$\square$\cr\kern0.07ex\hbox{\scalebox{0.9}{$\blacktriangle$}}}}}
\newcommand{\boxtriangle}{\mathrel{\ooalign{$\square$\cr\kern0.07ex\hbox{\scalebox{0.9}{$\triangle$}}}}}
\newcommand{\diamonddots}{\,\cdot\joinrel:\joinrel\cdot\,}

\newcommand{\righttherefore}{:\joinrel\cdot\,}

\newlist{todolist}{itemize}{2}
\setlist[todolist]{label=$\square$}

\usepackage[top=0.7cm,bottom=0.5cm,left=1cm,right=1cm]{geometry} % Seitenraender

\if\isdraft1
\fi

\title{
    Proportoids
}
\author{
    Christian Anti\'c
}
\address{
    christian.antic@icloud.com\\
    Vienna University of Technology\\
    Vienna, Austria
}

\begin{document}

\begin{abstract} 
	Analogical proportions are expressions of the form ``$a$ is to $b$ what $c$ is to $d$'' at the core of analogical reasoning, which itself is at the core of artificial intelligence. This paper contributes to the mathematical foundations of analogical proportions in the axiomatic tradition as initiated --- in the tradition of the ancient Greeks --- by Yves Lepage two decades ago. More precisely, we first introduce the name ``proportoid'' for sets endowed with a 4-ary analogical proportion relation satisfying a suitable set of axioms. We then study study different kinds of proportion-preserving mappings and relations and their properties. Formally, we define homomorphisms of proportoids as mappings $\mathsf H$ satisfying $a:b::c:d$ iff $\mathsf Ha: \mathsf Hb:: \mathsf Hc: \mathsf Hd$ for all elements and show that their kernel is a congruence. Moreover, we introduce (proportional) analogies as mappings $\mathsf A$ satisfying $a:b:: \mathsf Aa: \mathsf Ab$ for all elements $a$ and $b$ in the source domain and show how to compute partial analogies. We then introduce a number of useful relations between functions (including homomorphisms and analogies) on proportoids and study their properties. In a broader sense, this paper is a further step towards a mathematical theory of analogical proportions.
\end{abstract}

\maketitle


\tableofcontents

\section{Introduction}\label{§:I} 

Analogical proportions are expressions of the form ``$a$ is to $b$ what $c$ is to $d$'' at the core of analogical reasoning \cite{Hofstadter95a}, which itself is at the core of artificial intelligence \cite{Gust08,Hofstadter01}. The development of a formal theory of analogical proportions has recently gained some momentum \cite{Antic22,Antic22-2,Antic21-3,Antic23-4} with an immediate application to automatic logic programming by analogy in \cite{Antic23-23}. For a survey to computational approaches to analogical reasoning, we refer the reader to \cite{Hall89,Prade14}, and for an illustration of the usefulness of analogical proportions in AI to \cite{Prade21}.

The primary aim of the present paper is to establish a precise mathematical foundation for analogical proportions. In this sense, the work is intended as a first step: before developing concrete applications in the future, it is necessary to clarify the underlying structural properties of analogical proportions in a rigorous framework.

The purpose of this paper therefore is to further develop from a mathematical point of view the \textit{axiomatic} approach to analogical proportions as initiated --- in the tradition of the ancient Greeks --- by Lepage \cite{Lepage03} two decades ago. We will mostly remain agnostic regarding the concrete set of assumed axioms as we will only assume three evident properties. While individual proofs are syntactically short (which is often characteristic for the algebraic approach in which proofs are often modularly decomposed), their non-triviality lies in the choice of the underlying notions and the resulting structural perspective. The main contribution of the paper is not the complexity of individual derivations, but the identification of a coherent algebraic/relational framework in which the analogical proportion relation and its structure-preserving mappings and relations can be systematically analyzed. The selection of notions is guided by general algebraic principles, most importantly structure-preserving mappings and their induced relations. For example, the key notion of a ``proportional analogy'' to be introduced in \prettyref{§:p-Analogy} is inspired by the fundamental concept of a ``functor'' in category theory (see e.g. \cite{Awodey10}). That said, all concepts introduced in \prettyref{§:Hom}--\ref{§:p-Similarity} are original.

More formally, we define \textit{\textbf{proportoids}} \prettyref{§:Proportoids} as sets endowed with a 4-ary analogical proportion relation $a:b::c:d$ satisfying the axioms of reflexivity \prettyref{eq:r}, symmetry \prettyref{eq:s}, and inner symmetry \prettyref{eq:y}, and we define \textit{\textbf{subproportoids}}, \textit{\textbf{homomorphisms}} \prettyref{§:Hom}, and \textit{\textbf{congruences}} \prettyref{§:Con}. It is important to emphasize that all constructions in this paper are kept very general and do not refer to a concrete instance of an analogical proportion relation.\footnote{However, we do want to point out the algebro-logical framework in \cite{Antic22,Antic23-4}, which provides a \textit{canonical} analogical proportion relation in \textit{any} algebraic and logical structure containing functions and relations (see \prettyref{§:UA}) and has been applied to automatic logic programming by analogy in \cite{Antic23-23}.}

Functions preserving analogical proportions have already proven to be of practical interest (e.g. \cite{Couceiro17}), and studying their mathematical properties is essential for understanding proportions. We therefore define the notion of a \textit{\textbf{proportional homomorphism}} \prettyref{§:Hom} preserving the analogical proportion relation across different domains by satisfying a stronger version of the proportional inference principle of \cite{Couceiro17} which corresponds to the analogical jump in \cite{Davies87} (cf. \cite{Couceiro23}); cf. \prettyref{r:PPP}. In \prettyref{§:Con}, we then introduce the fundamental concept of a congruence on a proportoid and show in \prettyref{t:kernel} that the \textit{\textbf{kernel}} of every proportional homomorphism is a congruence as desired. 

In a similar vein, we then introduce the notion of a \textit{\textbf{proportional analogy}} \prettyref{§:p-Analogy} preserving the relationship between elements of the source domain. We show in \prettyref{t:A_sPPP} that analogies satisfy the \textit{\textbf{strong proportion-preserving property}} (cf. \prettyref{r:PPP}) given that the underlying proportoids are transitive. In \prettyref{§:PPA}, we provide a procedure for constructing \textit{\textbf{partial proportional analogies}} using an enumeration of the source structure and a selection function.

Functions between proportoids are fundamental. In \prettyref{§:PFR}, we therefore introduce a number of binary relations between functions on proportoids which we believe are appealing from a mathematical point of view since most of them are equivalence relations and some are congruences.

In \prettyref{§:PI}, we define the notion of a \textit{\textbf{proportional identity}} and show that it is a congruence relation in a wide class of proportoids (\prettyref{t:=_p_congruence}).

In \prettyref{§:Circles}, we define \textit{\textbf{proportional circles}} and show in \prettyref{t:C_cd} how they can be used to compute solutions to proportional equations from given ones.

In \prettyref{§:Function_proportions} we define analogical proportions between functions on proportoids point-wise and show some elementary properties. This provides a blueprint for how to lift results from proportoids to functions on proportoids.

In \prettyref{§:p-Similarity}, we introduce a notion of \textit{\textbf{similarity}} using analogical proportions and show in \prettyref{t:approx} that it is a congruence under mild conditions.

In \prettyref{§:UA}, we sketch relationships to universal algebra and predicate logic via the algebro-logical frameworks of analogical proportions in \cite{Antic22,Antic23-22,Antic23-4}.

Section \prettyref{§:Summary} provides a summary of the different axioms of an analogical proportion and its induced properties in a proportoid.

The paper then concludes with \prettyref{§:Conclusion} by sketching some potential lines of future research.

In a broader sense, this paper is a further step towards a mathematical theory of analogical proportions.

\section{Preliminaries}

A set $P$ is \textit{\textbf{denumerable}} iff there is a surjective mapping $\mathbb N\to P$. 

We denote the \textit{\textbf{identity function}} on any set by $\mathsf I$. 

A function $\mathsf F$ on a partially ordered set $(P,\leq)$ is \textit{\textbf{monotone}} iff $a\leq b$ implies $\mathsf Fa\leq \mathsf Fb$ for all $a,b\in P$. 

A \textit{\textbf{partial function}} from $P$ to $R$ is a function $\mathsf F:P\to R\cup\{\textbf{u}\}$ possibly yielding the value $\textbf{u}$ (``undefined'') for some elements of $P$. We define the \textit{\textbf{domain}} of such a function by
\begin{align*} 
    dom\, \mathsf F:=\{a\in P\mid \mathsf Fa\neq\textbf{u}\}.
\end{align*} We call $\mathsf F$ \textit{\textbf{total}} iff $dom\, \mathsf F=P$.

Let $\rho$ be a binary relation on unary mappings on $P$. An unary mapping $\mathsf F$ is \textit{\textbf{idempotent}} with respect to $\rho$ iff $\mathsf F\rho \mathsf F^2$. As usual, the relation $\rho$ is \textit{\textbf{left compatible}} iff
\begin{align*} 
    \mathsf F\rho \mathsf G \quad\Longrightarrow\quad \mathsf{EF}\rho \mathsf{EG},\quad\text{for all $\mathsf E$},
\end{align*} \textit{\textbf{right compatible}} iff
\begin{align*} 
    \mathsf F\rho \mathsf G \quad\Longrightarrow\quad \mathsf{FE}\rho \mathsf{GE},\quad\text{for all $\mathsf E$},
\end{align*} and \textit{\textbf{compatible}} iff
\begin{align*} 
    \mathsf E\rho \mathsf F \quad\text{and}\quad \mathsf G\rho \mathsf H \quad\Longrightarrow\quad \mathsf{EG}\rho \mathsf{FH}.
\end{align*} A left [right] compatible equivalence is called a \textit{\textbf{left}} [\textit{\textbf{right}}] \textit{\textbf{congruence}}. A \textit{\textbf{congruence}} is a compatible equivalence relation.

The following characterization of congruences is folklore (see e.g. \cite[Proposition 1.5.1]{Howie03}):

\begin{proposition}\label{p:left_right} A binary relation on a semigroup is a congruence iff it is both a left and a right congruence.
\end{proposition}

\section{Proportoids}\label{§:Proportoids}

In this section, we formally introduce proportoids as sets endowed with a 4-ary analogical proportion relation satisfying a minimal suitable set of axioms in the axiomatic tradition initiated by Yves Lepage \cite{Lepage03} two decades ago:%\footnote{The postulates listed here have mostly been introduced by \cite{Lepage03} and later used by many researchers.}

\begin{definition}\label{d:proportoids} A \textit{\textbf{proportoid}} is a pair $\mathfrak P= (P, ::)$ consisting of a non-empty set $P$ endowed with a 4-ary analogical proportion relation $::$ on $P$ satisfying the following axioms, for all $a,b,c,d\in P$:
\begin{align}
    \label{eq:r} &a:b:: a:b\quad\text{(reflexivity; r)},\\
    \label{eq:s} &a:b::c:d\quad\Longleftrightarrow\quad  c:d::a:b\quad\text{(symmetry; s)},\\
    \label{eq:y} &a:b::c:d \quad\Longleftrightarrow\quad  b:a::d:c\quad\text{(inner symmetry; y)}.
\end{align} 

Beyond that, we will consider the following properties, for $a,b,c,d,e,f\in P$ and injective function $f:P\to P$:
\begin{align}
    \label{eq:e} &a:a::c:c \quad\text{(inner reflexivity; e)},\\
    \label{eq:c} &a:b::c:d \quad\Longleftrightarrow\quad a:c::b:d \quad\text{(central permutation; c)},\\
    \label{eq:d} &a:a::a:d \quad\Longleftrightarrow\quad d=a \quad\text{(determinism; d)},\\
    \label{eq:o} &a:b::b:a \quad\text{(commutativity; o)},\\ % TODO wird nicht verwendet
    \label{eq:t} &a:b::c:d \quad\text{and}\quad c:d::e:f \quad\Longrightarrow\quad a:b::e:f\quad\text{(transitivity; t)},\\
    \label{eq:i} &a:b::c:d \quad\text{and}\quad b:e::d:f \quad\Longrightarrow\quad a:e::c:f\quad\text{(inner transitivity; i)},\\
    \label{eq:v} &a:a::c:d \quad\Longrightarrow\quad d=c\quad\text{(strong inner reflexivity; v)},\\
    \label{eq:x} &a:b::a:d \quad\Longrightarrow\quad d=b\quad\text{(strong reflexivity; x)},\\
    \label{eq:f} &a:fa::c:fc, \quad\text{(functionality; f)}.
\end{align} The symbol after the semicolon like e.g. ``t'' will be used in proofs as shortcuts for the respective axiom (in this case ``transitivity'') and to denote proportoids satisfying the axiom --- for example, a proportoid satisfying transitivity is called a \textit{\textbf{t-proportoid}} etc. Moreover, define the \textit{\textbf{solution set}} of the \textit{\textbf{proportional equation}} (or \textit{\textbf{p-equation}}) $a:b::c:x$ by
\begin{align*} 
    \mathscr S_{ \mathfrak P}(a:b::c:x) := \{d\in P \mid a:b::c:d\}.
\end{align*}
\end{definition}

\begin{remark}\label{r:Lepage03} Lepage \cite{Lepage03} uses the axioms \prettyref{eq:s}, \prettyref{eq:c}, \prettyref{eq:v}, and \prettyref{eq:x} in his axiomatization in the linguistic context.\footnote{Lepage \cite{Lepage03} used other names for the axioms --- we adapt here the terminology in \cite[§4.3]{Antic22}.} We agree with \prettyref{eq:s}. Although often accepted, central permutation \prettyref{eq:c} is debatable as shown by the simple counterexample:
\begin{center}
\begin{tikzpicture} 
    \node (a)               {$a$};
    \node (b) [above=of a]  {$b$};
    \node (c) [right=of a]  {$c$};
    \node (d) [right=of b]  {$d$};
    \draw (a) to (c);
\end{tikzpicture}
\end{center} Here we cannot say that $a$ is to $c$ as $b$ is to $d$ since there is a relation between $a$ and $c$ and none between $b$ and $d$. That central permutation is problematic has been observed by others as well (see e.g. \cite{Lim21,Murena18}). The remaining axioms in Lepage's original list are in general inadequate as well as demonstrated by simple counterexamples as in the proof of \cite[Theorem 28]{Antic22}. 
\end{remark}

\begin{example}\label{e:leq} Let $\leq$ be a partial ordering on $P$. Define
\begin{align*} 
    a:b::_\leq c:d \quad:\Longleftrightarrow\quad (a=b \quad\text{and}\quad c=d) \quad\text{or}\quad (a<b \quad\text{and}\quad c<d) \quad\text{or}\quad (a>b \quad\text{and}\quad d>c).
\end{align*} Then $(P, ::_\leq)$ is a proportoid.
\end{example}

\begin{example} Let $\theta$ be a reflexive and symmetric binary relation on $P$. Define
\begin{align*} 
    a:b::_\theta c:d \quad:\Longleftrightarrow\quad a\theta b \quad\text{and}\quad c\theta d.
\end{align*} Then $(P, ::_\theta)$ is a proportoid. If $\theta$ is transitive, then $(P, ::_\theta)$ is transitive.
\end{example}

\begin{example}\label{e:N} In $\mathbb N$, define the well-known \textit{\textbf{difference proportion relation}} (and see  \cite[Difference Proportion Theorem]{Antic22-2}) as follows:
\begin{align*} 
    a:b::c:d \quad:\Longleftrightarrow\quad a-b=c-d.
\end{align*} Then $(\mathbb N, ::)$ is a proportoid satisfying \textit{all} properties in \prettyref{d:proportoids}.
\end{example}

\begin{example}\label{e:M} A \textit{\textbf{metric space}} is a set $M$ together with a \textit{\textbf{distance function}} $\delta:M\times M\to \mathbb R$ satisyfing, for all $a,b,c\in M$,
\begin{align*} 
    &\delta(a,a) = 0,\\
    &\delta(a,b) > 0 \quad\text{whenever}\quad a\neq b \quad\text{\textit{\textbf{(positivity)}}},\\
    &\delta(a,b) = \delta(b,a) \quad\text{\textit{\textbf{(symmetry)}}},\\
    &\delta(a,c)\leq \delta(a,b) + \delta(b,c)\quad\text{\textit{\textbf{(triangle inequality)}}}.
\end{align*}

We endow every metric space $(M,\delta)$ with the 4-ary analogical proportion relation defined, for all $a,b,c,d\in M$, by
\begin{align}\label{eq:ab_delta_cd} 
    a : b ::_\delta c : d \quad:\Longleftrightarrow\quad \delta(a,b) = \delta(c,d).
\end{align} Then $(M, ::_\delta)$ is a proportoid.
\end{example}

\begin{example} Let $\circ$ be a binary operation on $P$. Define
\begin{align*} 
    a:b::_\circ c:d \quad:\Longleftrightarrow\quad a\circ b=c\circ d.
\end{align*} Then $(P, ::_\circ)$ is \textit{not} necessarily a proportoid. To see why, consider the operation $\circ$ defined, for all $a,b\in P$, by
\begin{align*} 
    a\circ b := b.
\end{align*} Let $c\in P$ be different from $a$. Then $a\circ b = c\circ b$, and therefore $a:b::c:b$. Then inner symmetry makes that $b:a::b:$ should hold. However, the latter fails to be the case because $b\circ a = a$ and $b\circ c = c$ and $a$ and $c$ are different by hypothesis. 
\end{example}

With a notion of structure there is always an associated notion of substructure:

\begin{definition} A proportoid $\mathfrak R= (R, ::)$ is a \textit{\textbf{subproportoid}} of $\mathfrak P= (P, ::)$ iff $R\subseteq P$ and $a:b::c:d$ holds in $\mathfrak R$ iff it holds in $\mathfrak P$, for all $a,b,c,d\in R$.
\end{definition}

Intuitively, a subproportoid preserves the analogical proportion relation between the elements which in general may not be the case. For example, it may be the case that $a:b::c:d$ holds in the proportoid $\mathfrak R$, whereas if we consider the larger proportoid $\mathfrak P$ we can find some $d'$ such that the relation between $a$ and $b$ and between $c$ and $d'$ is more similar than between $c$ and $d$.

\section{Proportional homomorphisms}\label{§:Hom}

With any kind of structure there comes a notion of a structure-preserving mapping and proportoids are no exception. In the rest of the paper, let $\mathfrak P= (P, ::_{ \mathfrak P})$ and $\mathfrak R = (R, ::_{ \mathfrak R})$ be proportoids where we often omit the indices from notation in case the underlying proportoids are clear from the context.

\begin{definition} We call a mapping $\mathsf H : P\to R$ a \textit{\textbf{(proportional) homomorphism}} iff for all $a,b,c,d\in P$,
\begin{align}\label{eq:hom} 
     a : b ::_{ \mathfrak P} c : d \quad\Longleftrightarrow\quad  \mathsf Ha: \mathsf Hb ::_{ \mathfrak R}\mathsf Hc : \mathsf Hd.
\end{align} A \textit{\textbf{(proportional) endomorphism}} is a homomorphism of the form $\mathfrak P\to \mathfrak P$, and a \textit{\textbf{(proportional) epimorphism}} is an onto homomorphism. A \textit{\textbf{(proportional) isomorphism}} is a bijective homomorphism. We call $\mathfrak P$ and $\mathfrak R$ are \textit{\textbf{(proportionally) isomorphic}} --- in symbols, $\mathfrak P\cong \mathfrak R$ --- iff there is a proportional isomorphism from $\mathfrak P$ to $\mathfrak R$.
\end{definition}

\begin{remark}\label{r:PPP} The only if part ``$\Longrightarrow$'' of the equivalence in \prettyref{eq:hom} is called the \textit{analogical inference principle} by \cite{Couceiro17} (and see \cite{Couceiro23}) and it can be viewed as a particular case of the so-called \textit{analogical jump} by \cite{Davies87}. We prefer using the term ``proportional'' instead of ``analogical'' and \textit{\textbf{proportion-preserving property}} (or \textit{\textbf{PPP}}) instead of ``analogical inference principle'', and we call the equivalence in \prettyref{eq:hom} the \textit{\textbf{strong proportion-preserving property}} (or \textit{\textbf{sPPP}}). 
\end{remark}

\prettyref{r:PPP} motivates the following definition:

\begin{definition} A \textit{\textbf{proportion-preserving mapping}} (or \textit{\textbf{pp-mapping}}) is any function $\mathsf P: \mathfrak P\to \mathfrak R$ satisfying 
\begin{align}\label{eq:pp} 
    a:b::_{ \mathfrak P}c:d \quad\Longrightarrow\quad \mathsf Pa: \mathsf Pb::_{ \mathfrak R} \mathsf Pc: \mathsf Pd,\quad\text{for all $a,b,c,d\in P$}.
\end{align}
\end{definition}

\begin{example} The \textit{\textbf{$k$-successor function}}, $k\geq 0$, given by
\begin{align*} 
    S^ka := a+k,
\end{align*} for all $a\in \mathbb N$, is a homomorphism on $(\mathbb N, ::)$, which is defined as in \prettyref{e:N}.
\end{example}

\begin{example}\label{e:B} Define the \textit{\textbf{boolean proportion relation}} in $\mathfrak R := (\{0,1\}, ::)$ as follows \cite{Antic21-3,Klein82}:
\begin{align*} 
    a:b::c:d \quad:\Longleftrightarrow\quad (a=b \quad\text{and}\quad c=d) \quad\text{or}\quad (a\neq b \quad\text{and}\quad c\neq d).
\end{align*} The negation function $\neg$ is an isomorphism since it is bijective and satisfies
\begin{align*} 
    a:b::c:d 
        \quad&\Longleftrightarrow\quad (a=b \quad\text{and}\quad c=d) \quad\text{or}\quad (a\neq b \quad\text{and}\quad c\neq d)\\
        \quad&\Longleftrightarrow\quad (\neg a=\neg b \quad\text{and}\quad \neg c=\neg d) \quad\text{or}\quad (\neg a\neq\neg b \quad\text{and}\quad \neg c\neq\neg d)\\
        \quad&\Longleftrightarrow\quad \neg a:\neg b::\neg c:\neg d.
\end{align*}
\end{example}

\begin{proposition} The space of all endomorphisms forms a monoid with respect to function composition with the neutral element given by the identity function.
\end{proposition}
\begin{proof} It follows from the definition that homomorphisms are closed under composition, that is, if $\mathsf H$ and $\mathsf G$ are homomorphisms, then $\mathsf H\circ G$ is a homomorphism as well:
\begin{align*} 
     a:b::c:d \quad&\Longleftrightarrow\quad  \mathsf Ha : \mathsf Hb :: \mathsf Hc : \mathsf Hd\\
        &\Longleftrightarrow\quad \mathsf G \mathsf Ha : \mathsf G\mathsf Hb :: \mathsf G\mathsf Hc : \mathsf G\mathsf Hd.
\end{align*} The identity function is clearly a homomorphism.
\end{proof}

\begin{theorem}[First Injectivity Theorem]\label{t:FIT} Every homomorphism defined on a d-proportoid is injective.
\end{theorem}
\begin{proof} The following derivation shows that $\mathsf Ha = \mathsf Hb$ implies $a=b$, for any homomorphism $\mathsf H :\mathfrak P\to \mathfrak R$ on a d-proportoid $\mathfrak P$ and $a,b\in P$:
\begin{prooftree}
    \AxiomC{$\mathsf Ha = \mathsf Hb$}
    \RightLabel{r}
    \UnaryInfC{$\mathsf Ha : \mathsf Ha :: \mathsf Ha : \mathsf Hb$}
    \RightLabel{sPPP \ref{r:PPP}}
    \UnaryInfC{$a:a::a:b$}
    \RightLabel{d}
    \UnaryInfC{$a=b$.}
\end{prooftree}
\end{proof}

\begin{remark} \prettyref{t:FIT} motivates the study of non-strong proportion-preserving functions as in \cite{Couceiro17} (cf. \prettyref{r:PPP}).
\end{remark}

\section{Proportional congruences}\label{§:Con}

In universal algebra, congruences provide a mechanism for factorizing algebras into equivalence classes compatible with the algebraic operations. Here, we require that the equivalence classes preserve the analogical proportion relation giving rise to the notion of a proportional congruence:\footnote{We will show in \prettyref{t:kernel} that proportional congruences and homomorphisms are connected via kernels.}

\begin{definition}\label{d:congruence} An equivalence relation $\theta$ on $P$ is a \textit{\textbf{(proportional) congruence}} on $\mathfrak P= (P, ::)$ iff for all elements $a,b,c,d,e,f,g,h\in P$,
\begin{prooftree}
    \AxiomC{$a\theta e\quad b\theta f\quad c\theta g\quad d\theta h$}
    \AxiomC{$a:b::c:d$}
    % \RightLabel{$\theta$}
    \BinaryInfC{$e:f::g:h$}
\end{prooftree} or, equivalently,
\begin{prooftree}
    \AxiomC{$a\theta e\quad b\theta f\quad c\theta g\quad d\theta h$}
    \UnaryInfC{$a:b::c:d \quad\Longleftrightarrow\quad  e:f::g:h$.}
\end{prooftree} %We denote the set of all congruences on $\mathfrak P$ by $p\text-Con(\mathfrak P)$.
\end{definition}

\begin{proposition} For any congruence $\theta$ on $\mathfrak P$,
\begin{prooftree}
    \AxiomC{$a\theta c$}
        \AxiomC{$b\theta d$}
        \RightLabel{.}
    \BinaryInfC{$a:b::c:d$}
\end{prooftree} 
\end{proposition}
\begin{proof} The following inference rule is an instance of the rule defining a congruence in \prettyref{d:congruence}:
\begin{prooftree}
    \AxiomC{$a\theta a\quad b\theta b\quad a\theta c\quad b\theta d$}
    \AxiomC{$a:b::a:b$}
    \BinaryInfC{$a:b::c:d$.}
\end{prooftree} Now observe that since $\theta$ is reflexive and analogical proportions are reflexive, we can omit $a\theta a$, $b\theta b$, and $a:b::a:b$ in the first line of the rule which immediately yields the first implication of the proposition.
\end{proof}

A standard construction in universal algebra is given by the kernel of a homomorphism (cf. \cite[Definition 6.7]{Burris00}) which we directly adapt here:

\begin{definition}\label{d:ker} The \textit{\textbf{kernel}} of a homomorphism $\mathsf H:\mathfrak P\to \mathfrak R$ is given by
\begin{align*} 
	ker\; \mathsf H := \left\{(a,b)\in P^2 \;\middle|\; \mathsf Ha = \mathsf Hb \right\}.
\end{align*}
\end{definition}

We now show that homomorphisms and congruences of proportoids are in the same way related as in universal algebra via kernels:

\begin{theorem}\label{t:kernel} The kernel of any homomorphism is a congruence.
\end{theorem}
\begin{proof} Let $\mathsf H: \mathfrak P\to \mathfrak R$ be a homomorphism. We show
\begin{prooftree}
    \AxiomC{$\mathsf Ha = He\quad \mathsf Hb = \mathsf Hf\quad \mathsf Hc = \mathsf Hg\quad \mathsf Hd = \mathsf Hh$}
    \AxiomC{$a:b::c:d$}
    \BinaryInfC{$e:f::g:h$}
\end{prooftree} for all $a,b,c,d,e,f,g,h\in P$, by the following derivation:
\begin{prooftree}
    \AxiomC{$\mathsf Ha = He\quad \mathsf Hb = \mathsf Hf\quad \mathsf Hc = \mathsf Hg\quad \mathsf Hd = \mathsf Hh$}
        \AxiomC{$a:b::c:d$}
        \RightLabel{PPP \ref{r:PPP}}
        \UnaryInfC{$ \mathsf Ha : \mathsf Hb :: \mathsf Hc : \mathsf Hd$}
    \BinaryInfC{$\mathsf He : \mathsf Hf :: \mathsf Hg : \mathsf Hh$}
    \RightLabel{sPPP \ref{r:PPP}}
    \UnaryInfC{$e:f::g:h$.}
\end{prooftree}
\end{proof}

\section{Proportional polymorphisms}\label{§:Poly}

The following definition follows the standard definition of a polymorphism on a relational proportoid:

\begin{definition} A \textit{\textbf{(proportional) polymorphism}} of $\mathfrak P= (P, ::)$ is any operation $f:P^n\to P$ preserving the analogical proportion relation in the sense that
\begin{prooftree}
    \AxiomC{$a_1:b_1::c_1:d_1$}
    \AxiomC{\ldots}
    \AxiomC{$a_n:b_n::c_n:d_n$}
    \RightLabel{PP}
    \TrinaryInfC{$ fa_1\ldots a_n:fb_1\ldots b_n::fc_1\ldots c_n:fd_1\ldots d_n$}
\end{prooftree} holds for all $a_i,b_i,c_i,d_i\in P$, $1\leq i\leq n$. %The set of all polymorphisms on $\mathfrak P$ is denoted by $p\text-Pol(\mathfrak P)$.
\end{definition}

\begin{remark} Notice that for any unary function $f:P\to P$, the above condition amounts to the proportion-preserving property (cf. \prettyref{r:PPP})
\begin{prooftree}
    \AxiomC{$a:b::c:d$}
    \RightLabel{PPP.}
    \UnaryInfC{$fa:fb::fc:fd$}
\end{prooftree} 
\end{remark}

\begin{definition} A \textit{\textbf{strong (proportional) polymorphism}} (or \textit{\textbf{s-polymorphism}}) is a polymorphism where the implication of the inference rule (PP) is turned into an equivalence.
\end{definition}

\begin{example}\label{e:sp} The iterated successor function $S^k$ is an s-polymorphism of $(\mathbb N, ::)$ defined as in \prettyref{e:N}, thus satisfying
\begin{align*} 
    a:b::c:d \quad\Longleftrightarrow\quad S^ka:S^kb::S^kc:S^kd,\quad\text{for all $a,b,c,d\in \mathbb N$ and $k\geq 0$}.
\end{align*} Moreover, addition is a polymorphism of $(\mathbb N, ::)$ thus satisfying
\begin{prooftree}
    \AxiomC{$a:b::c:d$}
    \AxiomC{$e:f::g:h$}
    \RightLabel{.}
    \BinaryInfC{$a+e:b+f::c+g:d+h$}
\end{prooftree}
\end{example}

Notice that we can use s-polymorphisms to compose and decompose proportions:

\begin{example} Since $S$ is an s-polymorphism of $(\mathbb N,S)$ as has been observed in \prettyref{e:sp}, we immediately obtain the following characterization of the analogical proportion relation in $(\mathbb N, ::)$:
\begin{align*} 
    a:b::c:d \quad\Longleftrightarrow\quad a-\min(a,b,c,d):b-\min(a,b,c,d) ::c-\min(a,b,c,d):d-\min(a,b,c,d).
\end{align*} For example, we can use the fact that $S$ is an s-polymorphism in the following way:
\begin{align*} 
    2:3::5:7 
        \quad&\Longleftrightarrow\quad SS0:SSS0::SSSSS0:SSSSSSS0\\
        &\Longleftrightarrow\quad S0:SS0::SSSS0:SSSSSS0\\
        &\Longleftrightarrow\quad 0:S0::SSS0:SSSSS0\\
        &\Longleftrightarrow\quad 0:1::3:5,
\end{align*} which fails in $(\mathbb N, ::)$ since $0-1\neq 3-5$. That is, we can decompose each analogical proportion $a:b::c:d$ in $(\mathbb N, ::)$ in such a way that at least one of $a,b,c,d$ is equal to 0.
\end{example}

\section{Proportional analogies}\label{§:p-Analogy}

In \prettyref{§:Hom}, we defined proportional homomorphisms as mappings satisfying the strong proportion-preserving property (cf. \prettyref{r:PPP}). In this section, we are interested in a related but different notion of proportion-preserving mapping defined as follows. First, we need the following construction (compare to \prettyref{d:proportoids}):

\begin{definition} Given two proportoids $\mathfrak P = (P, ::_{ \mathfrak P})$ and $\mathfrak R= (R, ::_{ \mathfrak R})$, we construct a \textit{\textbf{pair proportoid}} (or \textit{\textbf{pproportoid}}) $\mathfrak{PR} = (P, R, ::_{ \mathfrak{PR}})$, where $::_{ \mathfrak{PR}}\;\subseteq P^2\times R^2$ is a 4-ary analogical proportion relation on $P$ and $R$ satisfying, for all $a,b\in P\cap R$:
\begin{align} 
    a : b ::_{ \mathfrak{PR}} a : b \quad\text{(reflexivity; r)},
\end{align} and for all $a,b\in P$ and $c,d\in R$:
\begin{align} 
    &a : b ::_{ \mathfrak{PR}} c : d \quad\Longleftrightarrow\quad c : d ::_{ \mathfrak{RP}} a : b\quad\text{(symmetry; s)},\\
    &a : b ::_{ \mathfrak{PR}} c : d \quad\Longleftrightarrow\quad  b : a ::_{ \mathfrak{PR}} d : c\quad\text{(inner symmetry; y)}.
\end{align} Notice that every proportoid $(P, ::_{ \mathfrak P})$ can be turned into a pproportoid $(P, P, ::_{ \mathfrak{PP}})$ and we will not distinguish between the two. 

Moreover, we consider the following properties for all elements from the appropriate sets:
\begin{align*}
    &a : a ::_{ \mathfrak{PR}} c : c \quad\text{(inner reflexivity; e)},\\
    &a : b ::_{ \mathfrak{PR}} c : d\quad\Longleftrightarrow\quad a : c ::_{ \mathfrak{PR}} b : d \quad\text{(central permutation; c)},\\ 
    &a : a ::_{ \mathfrak{PR}} a : d \quad\Longleftrightarrow\quad d = a \quad\text{(determinism; d)},\\
    &a : b ::_{ \mathfrak{PR}} c : d \quad\text{and}\quad b : e ::_{ \mathfrak{PR}} d : f \quad\Longrightarrow\quad a : e ::_{ \mathfrak{PR}} c : f \quad\text{(inner transitivity; i)}.
\end{align*} Finally, let $\mathfrak Q = (Q, ::_{ \mathfrak Q})$ be an additional proportoid:
\begin{align} 
    \label{eq:ppt} a:b::_{ \mathfrak{PR}}c:d \quad\text{and}\quad c:d::_{ \mathfrak{RQ}} e:f \quad\Longrightarrow\quad a:b::_{ \mathfrak{PQ}} e:f\quad\text{(transitivity; t)}.
\end{align} As for proportoids, we add additional symbols to denote pproportoids satisfying specific properties; for example, an \textit{\textbf{i-pproportoid}} is a pproportoid satisfying inner transitivity. 

A \textit{\textbf{ppt-triple}} is a triple of pproportoids
\begin{align*} 
    \mathfrak{PRQ} = (P, R, ::_{ \mathfrak{PR}})(R,Q, ::_{ \mathfrak{RQ}})(P, Q, ::_{ \mathfrak{PQ}})
\end{align*} satisfying transitivity \prettyref{eq:ppt}. 

Finally, we define
\begin{align*} 
    \mathscr S_{ \mathfrak{PR}}(a:b::c:x) := \left\{d\in R \;\middle|\; a : b ::_{ \mathfrak{PR}} c : d\right\}.
\end{align*} 

We will often omit the indices from notation.
\end{definition}

\begin{definition} A \textit{\textbf{(proportional) analogy}} between the proportoids $\mathfrak P= (P, ::_{ \mathfrak P})$ and $\mathfrak R= (R, ::_{ \mathfrak R})$ is any mapping $\mathsf A:\mathfrak P\to \mathfrak R$ satisfying in a pproportoid $\mathfrak{PR}= (P,R, ::_{ \mathfrak{PR}})$ the properties
\begin{align*} 
    a:b::_{ \mathfrak{PR}}\mathsf Aa: \mathsf Ab, \quad\text{for all $a,b\in P$.}
\end{align*} We say that $\mathfrak P$ and $\mathfrak R$ are \textit{\textbf{(proportionally) analogous}} --- in symbols, $\mathfrak P:: \mathfrak R$ --- iff there is a bijective analogy from $\mathfrak P$ to $\mathfrak R$. %We say that $a\in P$ and $b\in R$ are \textit{\textbf{(proportionally) analogous}} --- in symbols, $a\approx_a b$ --- iff there is an analogy $\mathsf A: \mathfrak P\to \mathfrak R$ such that $b=\mathsf Aa$.
\end{definition}

\begin{fact} The identity function is an analogy in any e-proportoid satisfying inner reflexivity \prettyref{eq:e}.
\end{fact}

\begin{example}\label{e:S^k} The $k$-successor function $S^k$ is an analogy in $(\mathbb N, ::)$ defined as in \prettyref{e:N}.
\end{example}

\begin{example}\label{e:neg_analogy} The negation operation $\neg$ is an analogy on $(\{0,1\}, ::)$ defined as in \prettyref{e:B}.
\end{example}

\begin{example} The analogies in $(P, ::_\leq)$ as defined in \prettyref{e:leq} are the monotone functions on $P$.
\end{example}

\begin{example} The analogies on proportoids arising from metric spaces as in \prettyref{e:M} are distance-preserving mappings (i.e. isometries).
\end{example}

% \begin{definition} We call $\mathsf A:\mathfrak P\to \mathfrak P$ a \textit{\textbf{(proportional) permutable analogy}} (or \textit{\textbf{p-analogy}}) iff in addition to being an analogy it satisfies
% \begin{align*} 
%     a: \mathsf Aa::_{ \mathfrak P} b: \mathsf Ab,\quad\text{for all $a,b\in P$.}
% \end{align*} 
% \end{definition}

% \begin{remark} Of course, in every c-proportoid satisfying central permutation, analogies and p-analogies coincide.
% \end{remark}

\subsection{Strong proportion-preserving property}

Recall that homomorphisms satisfy the strong proportion-preserving property (cf. \prettyref{r:PPP})
\begin{align*} 
     a:b::_{ \mathfrak P} c:d \quad\Longleftrightarrow\quad  \mathsf Ha: \mathsf Hb::_{ \mathfrak R}\mathsf Hc: \mathsf Hd
\end{align*} for all $a,b,c,d\in P$. We have the following important result relating homomorphisms and analogies:

\begin{theorem}\label{t:A_sPPP} Every analogy on a ppt-triple $\mathfrak{PRP}$ satisfies the strong proportion-preserving property.
\end{theorem}
\begin{proof} Let $\mathsf A:\mathfrak P\to \mathfrak R$ be an analogy. We only prove the direction from left to right with the other direction being analogous: for any $a,b,c,d\in P$, we have
\begin{prooftree}
    \AxiomC{$a:b::_{ \mathfrak P} c:d$}
        \AxiomC{$ c:d::_{ \mathfrak{PR}} \mathsf Ac: \mathsf Ad$}
        \RightLabel{t}
    \BinaryInfC{$a:b::_{ \mathfrak{PR}} \mathsf Ac: \mathsf Ad$}
    \RightLabel{s}
    \UnaryInfC{$ \mathsf Ac: \mathsf Ad::_{ \mathfrak{RP}} a:b$}
        \AxiomC{$a:b::_{ \mathfrak{PR}} \mathsf Aa: \mathsf Ab$}
        \RightLabel{t}
    \BinaryInfC{$ \mathsf Ac: \mathsf Ad::_{ \mathfrak R}\mathsf Aa: \mathsf Ab$}
    \RightLabel{s}
    \UnaryInfC{$ \mathsf Aa: \mathsf Ab::_{ \mathfrak R}\mathsf Ac: \mathsf Ad$.}
\end{prooftree}
\end{proof}

\subsection{Second injectivity theorem}

Interestingly, the next result shows that analogies are injective in the wide range of dt-proportoids as an analogue to the First Injectivity \prettyref{t:FIT}:

\begin{theorem}[Second Injectivity Theorem]\label{t:SIT} Every analogy on a ppt-triple $\mathfrak{PRP}$, with $\mathfrak P$ a d-proportoid, is injective.
\end{theorem}
\begin{proof} Let $\mathsf A:\mathfrak P\to \mathfrak R$ be an analogy. We show that $\mathsf Aa=\mathsf Ab$ implies $a=b$, for any $a,b\in P$, by the following derivation:
\begin{prooftree}
    \AxiomC{$a:b:: \mathsf Aa: \mathsf Ab$}
    \RightLabel{$\mathsf Aa=\mathsf Ab$}
    \UnaryInfC{$a:b:: \mathsf Aa: \mathsf Aa$}
        \AxiomC{$ \mathsf Aa: \mathsf Aa::a:a$}
        \RightLabel{t}
    \BinaryInfC{$a:b::a:a$}
    \RightLabel{s}
    \UnaryInfC{$a:a::a:b$}
    \RightLabel{d}
    \UnaryInfC{$a=b$.}
\end{prooftree}
\end{proof}

\subsection{Closedness under composition}

The composition of two analogies yields another analogy given that the underlying proportoid is transitive which is shown in the next result:

\begin{theorem}\label{t:analogies_monoid} The space of all analogies on a t-proportoid forms a monoid with respect to function composition with the neutral element given by the identity function.
\end{theorem}
\begin{proof} First, it follows from the definition that analogies are closed under composition in case the underlying algebra is transitive, that is, if $\mathsf A: \mathfrak P\to \mathfrak P$ and $\mathsf B: \mathfrak P\to \mathfrak P$ are analogies, then $\mathsf B\circ \mathsf A$ is an analogy as well by the following derivation:
\begin{prooftree}
    \AxiomC{$a:b:: \mathsf Aa: \mathsf Ab$}
        \AxiomC{$ \mathsf Aa: \mathsf Ab:: \mathsf B\mathsf Aa: \mathsf B\mathsf Ab$}
        \RightLabel{t.}
    \BinaryInfC{$a:b:: \mathsf B\mathsf Aa: \mathsf B\mathsf Ab$}
\end{prooftree} The identity function is an analogy as an immediate consequence of reflexivity.
\end{proof}

\subsection{Proportional idempotency}

Every unary function on a proportoid can be applied repeatedly, which motivates the following definition:

\begin{definition} We say that $\mathsf F:\mathfrak P\to \mathfrak P$ is \textit{\textbf{(proportionally) idempotent}} iff
\begin{align*} 
     \mathsf Fa: \mathsf Fb :: \mathsf F\mathsf Fa: \mathsf F \mathsf Fb,\quad\text{holds for all $a,b\in P$.}
\end{align*}
\end{definition}

\begin{fact} Every analogy is idempotent.
\end{fact}

\begin{remark} Notice that by symmetry \prettyref{eq:s}, every idempotent function $\mathsf F:\mathfrak P\to \mathfrak P$ on a t-proportoid $\mathfrak P$ satisfies
\begin{align*} 
    \mathsf F^ma: \mathsf F^mb:: \mathsf F^na: \mathsf F^nb\quad\text{for all $m,n\geq 0$ and $a,b\in P$.}
\end{align*}
\end{remark}

\section{Partial proportional analogies}\label{§:PPA}

In this section, we provide a procedure for constructing \textit{partial} proportional analogies from a given enumeration of the source proportoid and selection functions on subsets of the target proportoid (which exist by the presumed axiom of choice). 

Concretely, let $\mathfrak{PR}$ be a denumerable pair of i-proportoids satisfying inner transitivity \prettyref{eq:i}, let $\sigma:2^R\to R\cup\{\textbf{u}\}$ be a selection function (recall that $\textbf{u}$ stands for ``undefined'') such that for every $S\subseteq R$,
\begin{align}
    \sigma S = \textbf{u} \quad:\Longleftrightarrow\quad  S = \emptyset,
\end{align} and let $e$ be an enumeration of $P$. We define
\begin{align*} 
    \mathscr S_{ \mathfrak{PR}}(a:b:: \textbf{u}:x) := \emptyset,
\end{align*} for all $a,b\in P$, that is, there can be no $d\in P$ such that $a:b:: \textbf{u}:d$.

We are now ready to introduce the main notion of this section:

\begin{definition}\label{d:A} Define the partial analogy $\mathsf A_{\sigma,e}: \mathfrak P\to \mathfrak R$, for a selection function $\sigma$ and enumeration $e$, inductively by
\begin{align*} 
    \mathsf A_{\sigma,e}e_1& := \sigma S,\\
    \mathsf A_{\sigma,e}e_{i+1}& :=\sigma \mathscr S_{ \mathfrak{PR}}(e_i:e_{i+1}:: \mathsf A_{\sigma,e}e_i:x),\quad i\geq 1.
\end{align*} Notice that the domain of $\mathsf A_{\sigma,e}$ cannot be empty since we always have $R\neq\emptyset$ by assumption and thus
\begin{align*} 
    e_1\in dom\,\mathsf A_{\sigma,e}.
\end{align*}
\end{definition}

The next result shows that the construction of \prettyref{d:A} always yields a partial analogy:

\begin{theorem}\label{t:analogy} On any i-pproportoid $\mathfrak{PR}$, the partial function $\mathsf A_{\sigma,e}: \mathfrak P\to \mathfrak R$ is an analogy on its non-empty domain, for every choice of $e$ and $\sigma$.
\end{theorem}
\begin{proof} We need to show
\begin{align}\label{eq:a_b_Aa_Ab} 
    a:b::_{ \mathfrak{PR}}\mathsf A_{\sigma,e}a: \mathsf A_{\sigma,e}b,\quad\text{for all $a,b\in dom\, \mathsf A_{\sigma,e}$.}
\end{align} By definition, we have
\begin{align*} 
    a=e_{i_a} \quad\text{and}\quad b=e_{i_b},\quad\text{for some $i_a,i_b\in\mathbb N$}.
\end{align*} Without loss of generality, we can assume $i_a\leq i_b$ since otherwise we can apply inner symmetry to obtain $b:a:: \mathsf A_{\sigma,e}b: \mathsf A_{\sigma,e}a$. So we have
\begin{align*} 
    i_b=i_a+j,\quad\text{for some $j\in\mathbb N$}.
\end{align*} By definition of $\mathsf A_{e,\sigma}$, we have
\begin{align*} 
    e_{i_a}:e_{i_a+1}&::_{ \mathfrak{PR}}\mathsf A_{\sigma,e}e_{i_a}: \mathsf A_{\sigma,e}e_{i_a+1}\\
    &\vdots\\
    e_{i_a+j-1}:e_{i_a+j}&::_{ \mathfrak{PR}}\mathsf A_{\sigma,e}e_{i_a+j-1}: \mathsf A_{\sigma,e}e_{i_a+j}
\end{align*} and
\begin{align*} 
    e_{i_a+k}\in dom\,\mathsf A_{\sigma,e},\quad\text{for all $1\leq k\leq j$}.
\end{align*} By the assumed inner transitivity \prettyref{eq:i} axiom, we obtain
\begin{align*} 
    e_{i_a}:e_{i_a+j}::_{ \mathfrak{PR}}\mathsf A_{\sigma,e}e_{i_a}: \mathsf A_{\sigma,e}e_{i_a+j}
\end{align*} which is equivalent to \prettyref{eq:a_b_Aa_Ab}.
\end{proof}

\begin{example} Let $e$ be the identity on $\mathbb N$, let $\sigma : 2^{ \mathbb N}\to\mathbb N\cup\{\textbf{u}\}$ be a selection function, and let $S$ be the unary successor function. Recall the difference proportion relation in $(\mathbb N, ::)$ of \prettyref{e:N} given by
\begin{align*} 
    a:b::c:d \quad:\Longleftrightarrow\quad  a-b = c-d.
\end{align*} \prettyref{d:A} yields (recall that we have chosen $e$ to be the identity function on $\mathbb N$ and it is thus omitted)
\begin{align*} 
    \mathsf A_\sigma 1 &= \sigma\mathbb N,\\
    \mathsf A_\sigma 2 &= \sigma\mathscr S_{(\mathbb N,S)}(1:2:: \mathsf A_\sigma 1:x)\\
        &= \sigma\{2-1+\mathsf A_\sigma 1\}\\
        &= 1+\mathsf A_{\sigma,e}1,\\
    \mathsf A_\sigma 3 &= \sigma \mathscr S_{(\mathbb N,S)}(2:3:: \mathsf A_\sigma 2:x)\\
        &= \sigma \mathscr S_{(\mathbb N,S)}(2:3::1+\mathsf A_\sigma 1:x)\\
        &= \sigma\{3-2+1+\mathsf A_\sigma 1\}\\
        &= 2+\mathsf A_\sigma 1,\\
        &\quad\vdots\\
    \mathsf A_\sigma i &= i+\mathsf A_\sigma 1,\quad\text{for all $i\in\mathbb N$.}
\end{align*} Since
\begin{align*} 
    dom\,\mathsf A_\sigma=\mathbb N,
\end{align*} the function $\mathsf A_\sigma:\mathbb{N\to N}$ is a total analogy.
\end{example}

% \begin{example}\label{e:abcd} Consider the pit-pair $\mathfrak{PR}$ where $\mathfrak P:= (\{a,b\},f^\mathfrak P)$and $\mathfrak R:= (\{c,d\},f^\mathfrak R)$are given respectively by
% \begin{center}
% \begin{tikzpicture}[node distance=2cm and 2cm] 
%   \node (a) {$a$};
%   \node (b) [above=of a] {$b$};
%   \node (c) [right=of a] {$c$};
%   \node (d) [above=of c] {$d$};
%   \draw[->] (a) to [edge label'={$f^\mathfrak P$}][loop] (a);
%   \draw[->] (b) to [edge label'={$f^\mathfrak P$}][loop] (b);
%   \draw[->] (c) to [edge label'={$f^\mathfrak R$}] (d);
%   \draw[->] (d) to [edge label'={$f^\mathfrak R$}][loop] (d);
% \end{tikzpicture}
% \end{center} Since
% \begin{align*} 
%   \mathfrak{PR}\not\models_A a:b::x:y,\quad\text{for all $x,y\in\{c,d\}$},
% \end{align*} there can be no analogy from $\mathfrak P$ to $\mathfrak R$.
% \end{example}

\begin{fact}\label{f:sPPP} Every partial analogy on a ppt-triple $\mathfrak{PRP}$ satisfies the strong proportion-preserving property on its domain.
\end{fact}
\begin{proof} Analogous to the proof of \prettyref{t:A_sPPP}.
\end{proof}

\prettyref{f:sPPP} tells us that our above procedure for constructing partial analogies automatically yields a procedure for constructing functions satisfying the strong proportion-preserving property and thus for constructing analogy-preserving functions as studied by \cite{Couceiro17} and \cite{Couceiro23}.

\begin{definition} We define the \textit{\textbf{cardinality}} of a partial analogy $\mathsf A_{\sigma,e}$ by
\begin{align*} 
     \# \mathsf A_{\sigma,e} :=
        \begin{cases}
            \min\left\{i\in\mathbb N \;\middle|\; \mathsf A_{\sigma,e}e_i = \textbf{u}\right\} - 1 & \text{$\mathsf A_{\sigma,e}$ is partial},\\
            \infty & \text{$\mathsf A_{\sigma,e}$ is total}.
        \end{cases}
\end{align*}
\end{definition}

\begin{definition}\label{d:maximal} We call $\mathsf A_{\sigma,e}$ \textit{\textbf{$\sigma$-maximal}} iff there is no analogy $\mathsf A_{\sigma',e}$ --- with respect to the \textit{same} enumeration $e$ --- such that
\begin{align*} 
    dom\,\mathsf A_{\sigma,e}\subsetneq dom\,\mathsf A_{\sigma',e}.
\end{align*} 
\end{definition}

\begin{fact}\label{f:e-maximal} $\mathsf A_{\sigma,e}$ is $\sigma$-maximal iff $\# \mathsf A_{\sigma,e}$ is maximal with respect to $\sigma$. Every total analogy $\mathsf A_{\sigma,e}$ is $\sigma$-maximal.
\end{fact}
\begin{proof} An immediate consequence of the fact that by construction,
\begin{align}\label{eq:AuAu} 
    \mathsf A_{\sigma,e}e_i=\textbf{u} \quad\text{implies}\quad \mathsf A_{\sigma,e}e_j=\textbf{u},\quad\text{for all $j\geq i$.}
\end{align}
\end{proof}

We shall now show that \textit{every} analogy is constructed from an enumeration and a selection function as above.

\begin{theorem}\label{t:A_sigma_e} For every partial analogy $\mathsf A:\mathfrak P\to \mathfrak R$ there is an enumeration $e$ of $P$ and a selection function $\sigma:2^R\to R\cup\{ \textbf{u}\}$ such that $\mathsf A=\mathsf A_{\sigma,e}$.
\end{theorem}
\begin{proof} Let $e$ be an enumeration of $P$ enumerating first the elements of $dom\,\mathsf A$ and then the rest in arbitrary order. Define, for every $i\geq 1$,
\begin{align*} 
    \mathsf A_{\sigma,e}e_1&:=\sigma R:=\mathsf Ae_1,\\
    \mathsf A_{\sigma,e}e_{i+1}&:=
        \begin{cases}
            \sigma\mathscr S_{ \mathfrak{PR}}(e_i:e_{i+1}:: \mathsf Ae_i:x):=\mathsf A e_{i+1} & e_{i+1}\in dom\,\mathsf A,\\
            \textbf{u} & \text{otherwise.}
        \end{cases}
\end{align*} We clearly have $\mathsf A_{\sigma,e}=\mathsf A$ by construction.
\end{proof}

Notice that $\sigma$-maximality is defined with respect to $\sigma$ for some fixed $e$ (see \prettyref{d:maximal}) --- this does \textit{not} gurantee that there is no analogy $\mathsf A$ constructed by other means with $dom\,\mathsf A_{\sigma,e}\subsetneq dom\,\mathsf A$. We say that $\mathsf A_{\sigma,e}$ is \textit{\textbf{maximal}} iff there is no such analogy $\mathsf A$. 

\begin{corollary} A partial analogy $\mathsf A_{\sigma,e}$ is maximal iff there is no analogy $\mathsf A_{\sigma',e'}$ such that $\# \mathsf A_{\sigma,e}<\#\mathsf A_{\sigma',e'}$.
\end{corollary}
\begin{proof} A direct consequence of \prettyref{f:e-maximal} and \prettyref{t:A_sigma_e}.
\end{proof}

Let $\mathsf A_{\sigma,e}$ be an arbitrary partial analogy and let $n$ be the cardinality of $\mathsf A_{\sigma,e}$. Suppose there is some $i\in\mathbb N$ such that
\begin{align*} 
    e_i\not\in dom\,\mathsf A_{\sigma,e}
\end{align*} satisfying
\begin{align}\label{eq:S_neq_emptyset} 
    \mathscr S(e_n:e_i:: \mathsf Ae_n:x)\neq\emptyset.
\end{align} Define the transposition $\tau:\mathbb{N\to N}$ by
\begin{align*} 
    \tau n&:=i\\
    \tau i&:=n\\
    \tau j&:=j,\quad\text{for all other $j\in\mathbb N$}.
\end{align*} We then have by \prettyref{eq:S_neq_emptyset}
\begin{align*} 
    \mathsf A_{\sigma,e_\tau}e_{\tau n}=\sigma \mathscr S(e_n:e_i:: \mathsf Ae_n:x)\neq\textbf{u},
\end{align*} which entails
\begin{align*} 
    \#\mathsf A_{\sigma,e_\tau}=n+1=\# \mathsf A_{\sigma,e}+1,
\end{align*} and thus
\begin{align*} 
    dom\,\mathsf A_{\sigma,e}\subsetneq dom\,\mathsf A_{\sigma,e_\tau}.
\end{align*} In other words, we have increased the cardinality of the partial analogy $\mathsf A_{\sigma,e}$ by one by transforming it into the partial analogy $\mathsf A_{\sigma,e_\tau}$. This procedure can be iterated until a (partial) analogy has been reached whose cardinality can no longer be increased.

\section{Proportional identity}\label{§:PI}

In this section, we shall introduce a proportional identity relation (cf. \prettyref{d:=_p}). First, we introduce an auxiliary identity relation:

\begin{definition}\label{d:=_c} Let $a,b\in P$. We define
\begin{align*} 
    a=_c b \quad:\Longleftrightarrow\quad a:b::c:c,\quad\text{for some $c\in P$.}
\end{align*}
\end{definition}

\begin{fact}
    \AxiomC{$a=b$}
    \RightLabel{.}
    \UnaryInfC{$a=_c b$}
    \DisplayProof
\end{fact}

\begin{proposition}\label{p:=_c} In any cet-proportoid, the relation $=_c$ is an equivalence relation for any element $c$.
\end{proposition} 
\begin{proof} The reflexivity and symmetry of $=_c$ follow from inner symmetry \prettyref{eq:y} and the assumed inner reflexivity \prettyref{eq:e} of the analogical proportion relation, respectively. To prove transitivity, we show the implication
\begin{prooftree}
    \AxiomC{$a =_c b$}
        \AxiomC{$b =_c d$}
    \BinaryInfC{$a =_c d$}
\end{prooftree} by the following derivation:
\begin{prooftree}
    \AxiomC{$a =_c b$}
    \UnaryInfC{$a:b::c:c$}
    \RightLabel{c}
    \UnaryInfC{$a:c::b:c$}
        \AxiomC{$b =_c d$}
        \UnaryInfC{$b:d::c:c$}
        \RightLabel{y}
        \UnaryInfC{$d:b::c:c$}
        \RightLabel{c}
        \UnaryInfC{$d:c::b:c$}
        \RightLabel{s}
        \UnaryInfC{$b:c::d:c$}
        \RightLabel{t}
    \BinaryInfC{$a:c::d:c$}
    \RightLabel{c}
    \UnaryInfC{$a:d::c:c$}
    \UnaryInfC{$a =_c d$.}
\end{prooftree}
\end{proof}

\begin{proposition} Every $n$-ary polymorphism $f$ satisfies, for any elements $a_1,b_1,\ldots,a_n,b_n\in P$,
\begin{prooftree}
    \AxiomC{$a_1 =_c b_1\quad\ldots\quad a_n =_c b_n$}
    \UnaryInfC{$fa_1\ldots a_n =_{fc\ldots c} fb_1\ldots b_n$.}
\end{prooftree}
\end{proposition}
\begin{proof}\hfill
\begin{prooftree}
    \AxiomC{$a_1 =_c b_1\quad\ldots\quad a_n =_c b_n$}
    \UnaryInfC{$a_1 : b_1 :: c : c\quad\ldots\quad a_n : b_n :: c : c$}
    \UnaryInfC{$fa_1\ldots a_n : fb_1\ldots b_n :: fc\ldots c : fc\ldots c$}
    \UnaryInfC{$fa_1\ldots a_n =_{fc\ldots c} fb_1\ldots b_n$.}
\end{prooftree}
\end{proof}

\begin{proposition}\label{p:e_f_g_h} In any ct-proportoid, we have for any elements $a,b,c,d,e,f,g,h,i\in P$,
\begin{prooftree}
    \AxiomC{$a =_i e\quad b =_i f\quad c =_i g\quad d =_i h$}
        \AxiomC{$a:b::c:d$}
        \RightLabel{ct.}
    \BinaryInfC{$e:f::g:h$}
\end{prooftree}
\end{proposition}
\begin{proof}\hfill
\begin{prooftree}
    \AxiomC{$c =_i g$}
    \UnaryInfC{$c:g::i:i$}
        \AxiomC{$d =_i h$}
        \UnaryInfC{$d:h::i:i$}
        \RightLabel{s}
        \UnaryInfC{$i:i::d:h$}
        \RightLabel{t}
        \BinaryInfC{$c:g::d:h$}
            \RightLabel{c}
            \UnaryInfC{$c:d::g:h$}
            \RightLabel{s}
            \UnaryInfC{$g:h::c:d$}
                \AxiomC{$a:b::c:d$}
                \RightLabel{s}
                \UnaryInfC{$c:d::a:b$}
                \RightLabel{t}
            \BinaryInfC{$g:h::a:b$}
                \AxiomC{$a =_i e$}
                \UnaryInfC{$a:e::i:i$}
                    \AxiomC{$b =_i f$}
                    \UnaryInfC{$b:f::i:i$}
                    \RightLabel{s}
                    \UnaryInfC{$i:i::b:f$}
                    \RightLabel{t}
                \BinaryInfC{$a:e::b:f$}
                \RightLabel{c}
                \UnaryInfC{$a:b::e:f$}
                \RightLabel{t}
            \BinaryInfC{$g:h::e:f$}
            \RightLabel{s}
            \UnaryInfC{$e:f::g:h$.}
\end{prooftree}
\end{proof}

We now generalize \prettyref{d:=_c} as follows:

\begin{definition}\label{d:=_p} Let $a,b\in P$. We define the \textit{\textbf{proportional identity relation}} (or \textit{\textbf{p-identity}}) by
\begin{align*} 
    a\stackrel{..}= b \quad:\Longleftrightarrow\quad a=_c b,\quad\text{for some $c\in P$,}
\end{align*} In case $a\stackrel{..}= b$, we call $a$ and $b$ \textit{\textbf{proportionally identical}} (or \textit{\textbf{p-identical}}).
\end{definition}

The following definition is analogous to the definition of a homomorphism kernel (cf. \prettyref{d:ker}):

\begin{definition} We define the \textit{\textbf{kernel}} of an analogy $\mathsf A: \mathfrak P\to \mathfrak R$ by
\begin{align*} 
    ker\; \mathsf A :=\left\{(a,b)\in P^2 \;\middle|\; \mathsf Aa= \mathsf Ab\right\}.
\end{align*}
\end{definition}

We have the following implications:

\begin{proposition}\label{p:ker_=_p}
    \AxiomC{$(a,b)\in ker\; \mathsf A$}
    \RightLabel{.}
    \UnaryInfC{$a\stackrel{..}= b$}
    \DisplayProof
\end{proposition}
\begin{proof} \hfill
\begin{prooftree}
    \AxiomC{$\mathsf A$ is an analogy}
    \UnaryInfC{$a:b:: \mathsf Aa: \mathsf Ab$}
        \AxiomC{$(a,b)\in ker\; \mathsf A$}
        \UnaryInfC{$\mathsf Aa= \mathsf Ab$}
    \BinaryInfC{$a:b:: \mathsf Aa: \mathsf Aa$}
    \UnaryInfC{$a\stackrel{..}= b$.}
\end{prooftree}
\end{proof}

\begin{proposition}\label{p:ab_in_ker} 
    \AxiomC{$(a,b)\in ker\; \mathsf A\qquad (c,d)\in ker\; \mathsf A$}
    \RightLabel{t.}
    \UnaryInfC{$a:b::c:d$}
    \DisplayProof
\end{proposition}
\begin{proof} \hfill
\begin{prooftree}
    \AxiomC{$(a,b)\in ker\; \mathsf A$}
    \RightLabel{\prettyref{p:ker_=_p}}
    \UnaryInfC{$a\stackrel{..}= b$}
        \AxiomC{$(c,d)\in ker\; \mathsf A$}
        \RightLabel{\prettyref{p:ker_=_p}}
        \UnaryInfC{$c\stackrel{..}= d$}
        \RightLabel{t, \prettyref{t:=_p->ap}} % todo forward ref
    \BinaryInfC{$a:b::c:d$.}
\end{prooftree}
\end{proof}

\begin{proposition}\label{p:=_p_=_c} In any et-proportoid, we have $a\stackrel{..}= b$ iff $a =_c b$ for all $c$.
\end{proposition}
\begin{proof} We only have to prove the direction from left to right. By definition of identity, the assumption $a\stackrel{..}= b$ implies that there is some $d$ such that $a:b::d:d$. By the assumed inner reflexivity \prettyref{eq:e}, we have $d:d::c:c$. Now apply transitivity to $a:b::d:d$ and $d:d::c:c$ to obtain $a:b::c:c$ which is equivalent to $a=_c b$.
\end{proof}

% \begin{fact} 
%   \AxiomC{$a=_c b$, for some $c$}
%   \UnaryInfC{$a\stackrel{..}= b$}
%   \DisplayProof holds in any proportoids $\mathfrak{PR}$.
% \end{fact}

\begin{proposition}\label{p:a=b->a=_pb} $a=b$ implies $a\stackrel{..}= b$ in any proportoid, and $a\stackrel{..}= b$ implies $a=b$ in any det-proportoid.
\end{proposition}
\begin{proof} The first implication is a direct consequence of inner reflexivity \prettyref{eq:e} which implies $a:a::c:c$, for every $c$. The second implication follows from:
\begin{align*} 
    a\stackrel{..}= b 
        & \quad\Longleftrightarrow\quad a:b::c:c, \quad\text{for some $c$},\\
        & \quad\stackrel{et}\Longleftrightarrow\quad a:b::c:c, \quad\text{for all $c$}, \quad\text{(\prettyref{p:=_p_=_c})}\\
        & \quad\Longleftrightarrow\quad a:b::a:a\\
        & \quad\stackrel d\Longleftrightarrow\quad a=b.
\end{align*}
\end{proof}

\begin{lemma}\label{l:a_b_e_e} 
    \AxiomC{$a:b::e:e$}
        \AxiomC{$c:d::f:f$}
        \RightLabel{t.}
    \BinaryInfC{$a:b::c:d$}
    \DisplayProof
\end{lemma}
\begin{proof}\hfill
\begin{prooftree}
    \AxiomC{$a:b::e:e$}
        \AxiomC{$e:e::f:f$}
            \AxiomC{$c:d::f:f$}
            \RightLabel{s}
            \UnaryInfC{$f:f::c:d$}
            \RightLabel{t}
        \BinaryInfC{$e:e::c:d$}
        \RightLabel{t}
    \BinaryInfC{$a:b::c:d$.}
\end{prooftree}
\end{proof}

\begin{theorem}\label{t:=_p_congruence} The proportional identity relation is a congruence in any cet-proportoid.
\end{theorem}
\begin{proof} That identity is an equivalence relation is an immediate consequence of Propositions \ref{p:=_c} and \ref{p:=_p_=_c} (where we had to assume the ``e'' \prettyref{eq:e} in ``cet'').

It remains to show:
\begin{prooftree}
    \AxiomC{$a\stackrel{..}= e\quad b\stackrel{..}= f\quad c\stackrel{..}= g\quad d\stackrel{..}= h$}
        \AxiomC{$a:b::c:d$}
    \BinaryInfC{$e:f::g:h$.}
\end{prooftree} To prove this implication, we proceed similar to the proof of \prettyref{p:e_f_g_h}:
\begin{prooftree}
    \AxiomC{$c\stackrel{..}= g$}
    \UnaryInfC{$c:g::i:i$}
        \AxiomC{$d\stackrel{..}= h$}
        \UnaryInfC{$d:h::j:j$}
        \RightLabel{\ref{l:a_b_e_e}}
    \BinaryInfC{$c:g::d:h$}
    \RightLabel{c}
    \UnaryInfC{$c:d::g:h$}
    \RightLabel{s}
    \UnaryInfC{$g:h::c:d$}
        \AxiomC{$a:b::c:d$}
        \RightLabel{s}
        \UnaryInfC{$c:d::a:b$}
        \RightLabel{t}
    \BinaryInfC{$g:h::a:b$}
        \AxiomC{$a\stackrel{..}= e$}
        \UnaryInfC{$a:e::k:k$}
            \AxiomC{$b\stackrel{..}= f$}
            \UnaryInfC{$b:f::l:l$}
            \RightLabel{\ref{l:a_b_e_e}}
        \BinaryInfC{$a:e::b:f$}
        \RightLabel{c}
        \UnaryInfC{$a:b::e:f$}
        \RightLabel{t}
    \BinaryInfC{$g:h::e:f$}
    \RightLabel{s}
    \UnaryInfC{$e:f::g:h$.}
\end{prooftree}
\end{proof}

\begin{theorem}\label{t:=_p->ap} 
    \AxiomC{$a\stackrel{..}= b$}
        \AxiomC{$c\stackrel{..}= d$}
        \RightLabel{t.}
    \BinaryInfC{$a:b::c:d$}
    \DisplayProof
\end{theorem}
\begin{proof}\hfill
\begin{prooftree}
    \AxiomC{$a\stackrel{..}= b$}
    \UnaryInfC{$a:b::e:e$}
        \AxiomC{$e:e::f:f$}
            \AxiomC{$c\stackrel{..}= d$}
            \UnaryInfC{$c:d::f:f$}
            \RightLabel{s}
            \UnaryInfC{$f:f::c:d$}
            \RightLabel{t}
        \BinaryInfC{$e:e::c:d$}
        \RightLabel{t}
    \BinaryInfC{$a:b::c:d$.}
\end{prooftree}
\end{proof}

\begin{proposition} Every proportional polymorphism $f:P^n\to P$ satisfies
\begin{prooftree}
    \AxiomC{$a_1\stackrel{..}= b_1\quad\ldots\quad a_n\stackrel{..}= b_n$}
    \UnaryInfC{$fa_1\ldots a_n\stackrel{..}= fb_1\ldots b_n$.}
\end{prooftree}
\end{proposition}
\begin{proof}\hfill
\begin{prooftree}
    \AxiomC{$a_1\stackrel{..}= b_1\quad\ldots\quad a_n\stackrel{..}= b_n$}
    \UnaryInfC{$a_1:b_1::c_1:c_1\quad\ldots\quad a_n:b_n::c_n:c_n$}
    \UnaryInfC{$fa_1\ldots a_n:fb_1\ldots b_n::fc_1\ldots c_n:fc_1\ldots c_n$}
    \UnaryInfC{$fa_1\ldots a_n\stackrel{..}= fb_1\ldots b_n$.}
\end{prooftree}
\end{proof}

\begin{theorem}\label{t:Afa=_pfAa} For any analogy $\mathsf A: \mathfrak P\to \mathfrak P$ on a cft-proportoid $\mathfrak P$ and any injective function $f:P\to P$,
\begin{align*} 
    \mathsf Afa\stackrel{..}= f \mathsf Aa,\quad\text{for all $a\in P$.}
\end{align*}
\end{theorem}
\begin{proof} Since $\mathsf A$ is an analogy by assumption, we have the following derivation, for all $a\in P$:
\begin{prooftree}
    \AxiomC{$f$ is injective}
    \RightLabel{f}
    \UnaryInfC{$\mathsf Aa:f\mathsf Aa::a:fa$}
        \AxiomC{$\mathsf A$ is an analogy}
        \UnaryInfC{$a:fa:: \mathsf Aa: \mathsf Afa$}
        \RightLabel{t}
    \BinaryInfC{$\mathsf Aa:f\mathsf Aa:: \mathsf Aa: \mathsf Afa$}
    \RightLabel{c}
    \UnaryInfC{$\mathsf Aa: \mathsf Aa:: \mathsf Afa:f \mathsf Aa$}
    \RightLabel{s}
    \UnaryInfC{$\mathsf Afa:f \mathsf Aa:: \mathsf Aa: \mathsf Aa$}
    \UnaryInfC{$\mathsf Afa=_{ \mathsf Aa} f \mathsf Aa$}
    \UnaryInfC{$\mathsf Afa\stackrel{..}= f \mathsf Aa$.}
\end{prooftree}
\end{proof}

% \begin{remark}\label{r:} Notice that in \prettyref{t:Afa=_pfAa}, the function $f$ is unary and in order to be able to reformulate the result for arbitrary functions $f:P^n\to P$, we have to consider analogical $(n,1)$-proportions of the form $a_1\ldots a_n:b::c_1\ldots c_n:d$ which are not the content of this paper.
% \end{remark}

\begin{proposition} 
    \AxiomC{$a\stackrel{..}= b$}
    \UnaryInfC{$\mathsf Pa\stackrel{..}= \mathsf Pb$}
    \DisplayProof holds for any pp-mapping $\mathsf P$.
\end{proposition}
\begin{proof}\hfill
\begin{prooftree}
    \AxiomC{$a\stackrel{..}= b$}
    \UnaryInfC{$a:b::c:c$\quad\text{for some $c$}}
    \RightLabel{\prettyref{eq:pp}}
    \UnaryInfC{$\mathsf Pa: \mathsf Pb:: \mathsf Pc: \mathsf Pc$\quad\text{for some $c$}}
    \UnaryInfC{$\mathsf Pa\stackrel{..}= \mathsf Pb$.}
\end{prooftree}
\end{proof}

\begin{proposition} For any homomorphism $\mathsf H: \mathfrak P\to \mathfrak R$,
\begin{align*} 
    a\stackrel{..}= b \quad\Longleftrightarrow\quad \mathsf Ha\stackrel{..}= \mathsf Hb.
\end{align*}
\end{proposition}
\begin{proof} We have
\begin{align*} 
    a\stackrel{..}= b 
        & \quad\Longleftrightarrow\quad a:b::c:c \quad\text{for some $c$}\\
        & \quad\Longleftrightarrow\quad \mathsf Ha: \mathsf Hb:: \mathsf Hc: \mathsf Hc \quad\text{for some $c$}\\
        & \quad\Longleftrightarrow\quad \mathsf Ha\stackrel{..}= \mathsf Hb.
\end{align*}
\end{proof}

\begin{definition} We call $a\in P$ a \textit{\textbf{p-fixed point}} of $\mathsf F:P\to P$ iff $a\stackrel{..}= \mathsf Fa$.
\end{definition}

\begin{proposition}\label{p:p-fixed} Let $\mathsf A$ be an analogy with a fixed point on a c-proportoid $(P, ::)$. Then every element of $P$ is a p-fixed point of $\mathsf A$.
\end{proposition}
\begin{proof} Let $a$ be an arbitrary element of $P$, and let $b\in P$ be a fixed point of $\mathsf A$, that is, $\mathsf Ab=b$. We then have
\begin{prooftree}
    \AxiomC{$\mathsf A$ is an analogy}
    \UnaryInfC{$a:b:: \mathsf Aa: \mathsf Ab$}
    \RightLabel{c}
    \UnaryInfC{$a: \mathsf Aa::b: \mathsf Ab$}
    \RightLabel{$\mathsf Ab=b$}
    \UnaryInfC{$a: \mathsf Aa::b:b$}
    \UnaryInfC{$a\stackrel{..}= Aa$.}
\end{prooftree}
\end{proof}

\section{Proportional function relations}\label{§:PFR}

We now turn our attention to functions on proportoids where we wish to be able to compare two functions with respect to the analogical proportion relation.

\subsection{Functional proportionality}\label{§:FP}

Given $\mathsf F:\mathfrak P\to \mathfrak R$ and $\mathsf G: \mathfrak P\to \mathfrak Q$, define their \textit{\textbf{functional proportionality relation}} by
\begin{align*} 
    \mathsf F:: \mathsf G \quad:\Longleftrightarrow\quad \mathsf Fa: \mathsf Fb::_{ \mathfrak{RQ}} \mathsf Ga: \mathsf Gb,\quad\text{for all $a,b\in P$}.
\end{align*} In case $\mathsf F:: \mathsf G$, we say that $\mathsf F$ and $\mathsf G$ are \textit{\textbf{proportional}}. This can be depicted as follows:
\begin{center}
\begin{tikzpicture} 
    \node (Fa) {$\mathsf Fa$};
    \node (Fb) [below of=Fa,yshift=-0.5cm] {$\mathsf Fb$};
    \node (a) [right=of Fa] {$a$};
    \node (b) [below=of a,yshift=-0.03cm] {$b$};
    \node (Ga) [right=of a] {$\mathsf Ga$};
    \node (Gb) [below=of Ga,yshift=0.03cm] {$\mathsf Gb$};
    \draw[->,dashed] (a) to (Fa);
    \draw[->,dashed] (b) to (Fb);
    \draw[->,dashed] (a) to (Ga);
    \draw[->,dashed] (b) to (Gb);
    \draw (Fa) to (Fb);
    \draw (Ga) to (Gb);
\end{tikzpicture}
\end{center} In case $\mathsf A$ and $\mathsf B$ are analogies, the figure for $\mathsf A:: \mathsf B$ can be refined to:
\begin{center}
\begin{tikzpicture} 
    \node (Aa) {$\mathsf Aa$};
    \node (Ab) [below of=Aa,yshift=-0.5cm] {$\mathsf Ab$};
    \node (a) [right=of Aa] {$a$};
    \node (b) [below=of a,yshift=-0.03cm] {$b$};
    \node (Ba) [right=of a] {$\mathsf Ba$};
    \node (Bb) [below=of Ga] {$\mathsf Bb$};
    \draw[->,dashed] (a) to (Aa);
    \draw[->,dashed] (b) to (Ab);
    \draw[->,dashed] (a) to (Ba);
    \draw[->,dashed] (b) to (Bb);
    \draw (a) to (b);
    \draw (Aa) to (Ab);
    \draw (Ba) to (Bb);
\end{tikzpicture}
\end{center}

\begin{fact} $\mathsf F:: \mathsf I$ iff $\mathsf F$ is an analogy.
\end{fact}

Before we show that functional proportionality is a congruence relation, we shall first prove an auxiliary lemma which is interesting in its own right as it shows that functional proportionality is in a sense compatible with analogical proportions.

\begin{lemma}\label{l:FP} For any analogies $\mathsf A: \mathfrak P\to \mathfrak R$ and $\mathsf B: \mathfrak P\to \mathfrak Q$ on a ppt-triple $\mathfrak{QQR}$,
\begin{prooftree}
    \AxiomC{$a:b::_{ \mathfrak P} c:d$}
        \AxiomC{$\mathsf Aa: \mathsf Ab::_{ \mathfrak{RQ}} \mathsf Ba: \mathsf Bb$}
        \RightLabel{t.}
    \BinaryInfC{$\mathsf Aa: \mathsf Ab::_{ \mathfrak{RQ}} \mathsf Bc: \mathsf Bd$}
\end{prooftree}
\end{lemma}
\begin{proof}\hfill
\begin{center}
    \AxiomC{$a:b::_{ \mathfrak P} c:d$}
    \RightLabel{sPPP \ref{t:A_sPPP}}
    \UnaryInfC{$\mathsf Ba: \mathsf Bb ::_{ \mathfrak Q} \mathsf Bc: \mathsf Bd$}
    \RightLabel{s}
    \UnaryInfC{$\mathsf Bc: \mathsf Bd ::_{ \mathfrak Q} \mathsf Ba: \mathsf Bb$}
        \AxiomC{$\mathsf Aa: \mathsf Ab ::_{ \mathfrak{RQ}} \mathsf Ba: \mathsf Bb$}
        \RightLabel{s}
        \UnaryInfC{$\mathsf Ba: \mathsf Bb::_{ \mathfrak{QR}} \mathsf Aa: \mathsf Ab$}
        \RightLabel{t}
    \BinaryInfC{$\mathsf Bc: \mathsf Bd::_{ \mathfrak{QR}} \mathsf Aa: \mathsf Ab$}
    \RightLabel{s}
    \UnaryInfC{$\mathsf Aa: \mathsf Ab::_{ \mathfrak{RQ}} \mathsf Bc: \mathsf Bd$.}
    \DisplayProof
\end{center}
\end{proof}

\begin{theorem}\label{t:fp_cong} Functional proportionality of analogies is a congruence on any t-proportoid.
\end{theorem}
\begin{proof} Reflexivity, symmetry, and transitivity of the proportionality relation follows by reflexivity, symmetry, and the assumed transitivity of analogical proportions. It remains to show that it is compatible with composition:
\begin{prooftree}
    \AxiomC{$\mathsf A:: \mathsf B$}
    \UnaryInfC{$\mathsf Aa: \mathsf Ab:: \mathsf Ba: \mathsf Bb$}
        \AxiomC{$\mathsf C:: \mathsf D$}
        \UnaryInfC{$\mathsf Ca: \mathsf Cb:: \mathsf Da: \mathsf Db$}
        \RightLabel{t, \ref{l:FP}}
    \BinaryInfC{$\mathsf{AC}a: \mathsf{AC}b:: \mathsf{BD}a: \mathsf{BD}b$}
    \UnaryInfC{$\mathsf{AC}:: \mathsf{BD}$.}
\end{prooftree}
\end{proof}

\begin{example} All generalized successor functions $S^k$ and $S^\ell$ are functionally proportional in $(\mathbb N, ::)$ (cf. \prettyref{e:N}), that is, $S^k :: S^\ell$ holds for all $k,\ell\geq 0$.
\end{example}

\begin{proposition} Functional proportionality is reflexive and symmetric on any proportoid. If the underlying proportoid is transitive, then functional proportionality is transitive and therefore an equivalence relation.
\end{proposition}
\begin{proof} Reflexivity and symmetry follow from the reflexivity and symmetry of analogical proportions, and the assumed transitivity induces transitivity.
\end{proof}

% \begin{fact} $\mathsf F :: G$ iff $\mathsf F\circ H :: G\circ H$ for all $\mathsf H$.
% \end{fact}
% \begin{proof} The direction from right to left holds trivially by choosing $\mathsf H$ to be the identity function. The direction from left to right is an immediate consequence of the definition.
% \end{proof}

\begin{theorem} Let $\mathfrak{PRR}$ be a ppt-triple, let $\mathsf A: \mathfrak P\to \mathfrak R$ be an analogy, and let $\mathsf F: \mathfrak P\to \mathfrak R$ be an arbitrary function. If $\mathsf A:: \mathsf F$, then $\mathsf F$ is an analogy.
\end{theorem}
\begin{proof}\hfill
\begin{prooftree}
    \AxiomC{$\mathsf A$ is an analogy}
    \UnaryInfC{$a:b::_{ \mathfrak{PR}}\mathsf Aa: \mathsf Ab$}
        \AxiomC{$\mathsf A:: \mathsf F$}
        \UnaryInfC{$\mathsf Aa: \mathsf Ab::_{ \mathfrak R} \mathsf Fa: \mathsf Fb$}
        \RightLabel{t}
    \BinaryInfC{$a:b::_{ \mathfrak{PR}} \mathsf Fa: \mathsf Fb$.}
\end{prooftree}
\end{proof}

\begin{theorem}\label{t:A_B} All analogies $\mathsf{A,B}: \mathfrak P\to \mathfrak R$ on ppt-triples $\mathfrak{RPR}$ are functionally proportional.
\end{theorem}
\begin{proof} We have the following derivation, for any $a,b\in P$:
\begin{prooftree}
    \AxiomC{$a:b::_{ \mathfrak{PR}} \mathsf Aa: \mathsf Ab$}
    \RightLabel{s}
    \UnaryInfC{$ \mathsf Aa: \mathsf Ab::_{ \mathfrak{RP}} a:b$}
        \AxiomC{$a:b::_{ \mathfrak{PR}} \mathsf Ba: \mathsf Bb$}
        \RightLabel{t}
    \BinaryInfC{$ \mathsf Aa: \mathsf Ab::_{ \mathfrak R} \mathsf Ba: \mathsf Bb$}
    \UnaryInfC{$\mathsf A :: \mathsf B$.}
\end{prooftree}
\end{proof}

The next result connects analogies with homomorphisms with respect to functional proportionality:

\begin{theorem} For any analogy $\mathsf A$ and homomorphism $\mathsf H$ on a t-proportoid, we have $$\mathsf{HA}:: \mathsf{AH}.$$
\end{theorem}
\begin{proof}\hfill
\begin{prooftree}
    \AxiomC{$\mathsf A$ is an analogy}
    \UnaryInfC{$a:b:: \mathsf Aa: \mathsf Ab$}
    \RightLabel{$\mathsf H$ is a homomorphism}
    \UnaryInfC{$\mathsf Ha: \mathsf Hb:: \mathsf{HA}a: \mathsf{HA}b$}
    \RightLabel{s}
    \UnaryInfC{$\mathsf{HA}a: \mathsf{HA}b:: \mathsf Ha: \mathsf Hb$}
        \AxiomC{$\mathsf A$ is an analogy}
        \UnaryInfC{$\mathsf Ha: \mathsf Hb:: \mathsf{AH}: \mathsf{AH}b$}
        \RightLabel{t}
    \BinaryInfC{$\mathsf{HA}a: \mathsf{HA}b:: \mathsf{AH}a: \mathsf{AH}b$}
    \UnaryInfC{$\mathsf{HA}:: \mathsf{AH}$}
\end{prooftree}
\end{proof}

\subsection{Diamond equivalence}

Given $\mathsf F, \mathsf G:\mathfrak P\to \mathfrak R$, define their (\textit{\textbf{proportional}}) \textit{\textbf{diamond equivalence}} by
\begin{align*} 
    \mathsf F\diamonddots \mathsf G \quad:\Longleftrightarrow\quad \mathsf Fa: \mathsf Ga::_{ \mathfrak R} \mathsf Fb: \mathsf Gb,\quad\text{for all $a,b\in P$}.
\end{align*} This can be depicted as follows:
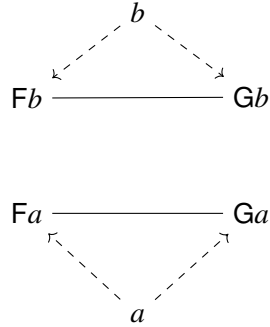
\begin{center}
\begin{tikzpicture} 
    \node (a) {$a$};
    \node (a') [above=of a] {};
    \node (Fa) [left=of a'] {$\mathsf Fa$};
    \node (Ga) [right=of a'] {$\mathsf Ga$};
    \node (b') [above=of a'] {};
    \node (b) [above=of b'] {$b$};
    \node (Fb) [above=of Fa] {$\mathsf Fb$};
    \node (Gb) [above=of Ga] {$\mathsf Gb$};
    \draw[->,dashed] (a) to (Fa);
    \draw[->,dashed] (a) to (Ga);
    \draw (Fa) to (Ga);
    \draw[->,dashed] (b) to (Fb);
    \draw[->,dashed] (b) to (Gb);
    \draw (Fb) to (Gb);
\end{tikzpicture}
\end{center} Notice the similarity to functional proportionality in \prettyref{§:FP}.

\begin{proposition}\label{p:diamond_equ_i} All analogies are diamond equivalent on i-proportoids.
\end{proposition}
\begin{proof}\hfill
\begin{prooftree}
    \AxiomC{$\mathsf A$ is analogy}
    \UnaryInfC{$a: \mathsf Aa::b: \mathsf Ab$}
    \RightLabel{y}
    \UnaryInfC{$\mathsf Aa:a:: \mathsf Ab:b$}
        \AxiomC{$\mathsf B$ is analogy}
        \UnaryInfC{$a: \mathsf Ba::b: \mathsf Bb$}
    \RightLabel{i}
    \BinaryInfC{$\mathsf Aa: \mathsf Ba:: \mathsf Ab: \mathsf Bb$}
    \UnaryInfC{$\mathsf A\diamonddots \mathsf B$.}
\end{prooftree}
\end{proof}

\begin{fact} $\mathsf A\diamonddots \mathsf I$ iff $\mathsf A$ is an analogy.
\end{fact}

\begin{proposition}\label{p:diamond_equ} Diamond equivalence is an equivalence relation on i-proportoids.
\end{proposition}
\begin{proof} Reflexivity of diamond follows from inner reflexivity of analogical proportions, and symmetry holds trivially. The following derivation proves transitivity of diamond:
\begin{prooftree}
    \AxiomC{$\mathsf F\diamonddots G$}
    \UnaryInfC{$\mathsf Fa: \mathsf Ga:: \mathsf Fb: \mathsf Gb$}
       \AxiomC{$G\diamonddots \mathsf H$}
       \UnaryInfC{$\mathsf Ga: \mathsf Ha:: \mathsf Gb: \mathsf Hb$}
       \RightLabel{i}
    \BinaryInfC{$\mathsf Fa: \mathsf Ha:: \mathsf Fb: \mathsf Hb$}
    \UnaryInfC{$\mathsf F\diamonddots \mathsf H$.}
\end{prooftree}
\end{proof}

\begin{fact}\label{f:diamond_fp} In any c-proportoid, we have $\mathsf F\diamonddots G$ iff $\mathsf F:: \mathsf G$.
\end{fact}

\begin{fact}\label{f:diamond_con} Diamond equivalence is a congruence on ct-proportoids.
\todo[inline]{Wie passt das mit \prettyref{p:diamond_equ} zusammen, wo i \prettyref{eq:i} verlangt wird? Ist nicht jede Con auch eine Equ?}
\end{fact}
\begin{proof} A direct consequence of \prettyref{t:fp_cong} and \prettyref{f:diamond_fp}.
\end{proof}

\begin{example} In $(\mathbb N, ::)$ defined as in \prettyref{e:N}, we have
\begin{align*} 
    \mathsf F\diamonddots \mathsf G \quad\Longleftrightarrow\quad \mathsf Fa- \mathsf Fb= \mathsf Ga- \mathsf Gb,\quad\text{for all $a,b\in\mathbb N$}.
\end{align*} In particular, we have $S^k\diamonddots S^\ell$, for all $k,\ell\geq 0$.
\end{example}

\subsection{Equivalence}\label{§:Equivalence}

Given mappings $\mathsf F, \mathsf G:\mathfrak P\to \mathfrak R$, define their (\textit{\textbf{proportional}}) \textit{\textbf{equivalence}} by
\begin{align*} 
    \mathsf F\equiv \mathsf G \quad:\Longleftrightarrow\quad a:a::_{ \mathfrak{PR}} \mathsf Fa: \mathsf Ga,\quad\text{for all $a\in P$}.
\end{align*} This situation can be depicted as follows:
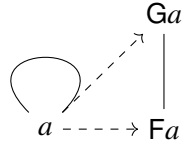
\begin{center}
\begin{tikzpicture} 
    \node (a) {$a$};
    \node (Fa) [right=of a] {$\mathsf Fa$};
    \node (Ga) [above=of Fa] {$\mathsf Ga$};
    \draw (a) to [loop] (a);
    \draw[->,dashed] (a) to (Fa);
    \draw[->,dashed] (a) to (Ga);
    \draw (Fa) to (Ga);
\end{tikzpicture}
\end{center}

The following result provides a simple way to show that two mappings are \textit{not} equivalent given that determinism \prettyref{eq:d} holds:

\begin{theorem}\label{t:equiv_fix} Proportionally equivalent mappings have the same fixed points in d-proportoids.
\end{theorem}
\begin{proof} Let $a$ be a fixed point of $\mathsf F$ thus satisfying $\mathsf Fa=a$. Then, by determinism \prettyref{eq:d} and since $\mathsf F$ and $\mathsf G$ are equivalent by assumption, we have
\begin{align*} 
    a:a:: \mathsf Fa: \mathsf Ga \quad\Longleftrightarrow\quad \mathsf Ga=a,
\end{align*} which means that $a$ is a fixed point of $\mathsf G$ as well. By inner reflexivity, an analogous argument shows that every fixed point of $\mathsf G$ is a fixed of $\mathsf F$.
\end{proof}

\begin{corollary}\label{c:F_equiv_I} $\mathsf F\equiv \mathsf I$ iff $\mathsf F = \mathsf I$ holds in d-proportoids.
\end{corollary}
\begin{proof} A direct consequence of \prettyref{t:equiv_fix}.
\end{proof}

\begin{theorem}\label{t:equiv_cong} Proportional equivalence is an equivalence relation on it-proportoids, and it is a congruence for all analogies on t-proportoids.
\end{theorem}
\begin{proof} Reflexivity of proportional equivalence follows from \prettyref{eq:r}, symmetry follows from \prettyref{eq:y}, and transitivity follows from \prettyref{eq:i}. 

It remains to show that equivalence is compatible with composition of analogies. In case $\mathsf A$ is an analogy, we have
\begin{prooftree}
    \AxiomC{$\mathsf B\equiv \mathsf C$}
    \UnaryInfC{$a:a:: \mathsf Ba: \mathsf Ca$}
    \RightLabel{$\mathsf A$ is an analogy}
    \UnaryInfC{$a:a:: \mathsf{AB}a: \mathsf{AC}a$}
    \UnaryInfC{$\mathsf{AB}\equiv \mathsf{AC}$.}
\end{prooftree} Moreover, we always have (even if $\mathsf A$ is not an analogy)
\begin{prooftree}
    \AxiomC{inner reflexivity}
    \UnaryInfC{$a:a:: \mathsf Aa: \mathsf Aa$}
       \AxiomC{$\mathsf B\equiv \mathsf C$}
       \UnaryInfC{$\mathsf Aa: \mathsf Aa:: \mathsf{BA}a: \mathsf{CA}a$}
       \RightLabel{t}
    \BinaryInfC{$a:a:: \mathsf{BA}a: \mathsf{CA}a$}
    \UnaryInfC{$\mathsf{BA}\equiv \mathsf{CA}$.}
\end{prooftree} This shows that equivalence is left and right compatible, which by \prettyref{p:left_right} means that it is a congruence.
\end{proof}

\begin{theorem} Proportional equivalence is left cancellative for analogies on any t-proportoid in the strong sense that for all analogies $\mathsf{A,B,C}$,
\begin{align*} 
    \mathsf{AB}\equiv \mathsf{AC} \quad\Longleftrightarrow\quad \mathsf B\equiv \mathsf C.
\end{align*} In case $\mathsf A$ commutes with $\mathsf B$ and $\mathsf C$ in the sense that
\begin{align}\label{eq:E_commutes_with_F_and_G} 
    \mathsf{BA}a: \mathsf{CA}a:: \mathsf{AB}a: \mathsf{AC}a,\quad\text{for all $a\in P$},
\end{align} it follows that equivalence is right cancellative in the strong sense as well, that is,
\begin{align*} 
    \mathsf{BA}\equiv \mathsf{CA} \quad\Longleftrightarrow\quad \mathsf A\equiv \mathsf C.
\end{align*}
\end{theorem}
\begin{proof} For the direction from left to right, we have the following derivation:
\begin{prooftree}
    \AxiomC{$\mathsf{AB}\equiv \mathsf{AC}$}
    \UnaryInfC{$a:a:: \mathsf{AB}a: \mathsf{AC}a$}
       \AxiomC{$\mathsf A$ is an analogy}
       \UnaryInfC{$\mathsf{AB}a: \mathsf{AC}a:: \mathsf Ba: \mathsf Ca$}
       \RightLabel{t}
    \BinaryInfC{$a:a:: \mathsf Ba: \mathsf Ca$}
    \UnaryInfC{$\mathsf B\equiv \mathsf C$.}
\end{prooftree} The direction from right to left holds trivially:
\begin{prooftree}
    \AxiomC{$\mathsf B\equiv \mathsf C$}
    \UnaryInfC{$a:a:: \mathsf Ba: \mathsf Ca$}
    \RightLabel{$\mathsf A$ is an analogy}
    \UnaryInfC{$a:a:: \mathsf{AB}a: \mathsf{AC}a$}
    \UnaryInfC{$\mathsf{AB}\equiv \mathsf{AC}$.}
\end{prooftree}

For the second part, we assume that $\mathsf A$ commutes with $\mathsf B$ and $\mathsf C$ in the sense of \prettyref{eq:E_commutes_with_F_and_G}. The direction from left to right is shown by the following derivation:
\begin{prooftree}
    \AxiomC{$\mathsf{BA}\equiv \mathsf{CA}$}
    \UnaryInfC{$a:a:: \mathsf{BA}a: \mathsf{CA}a$}
        \AxiomC{$\mathsf A$ commutes with $\mathsf B$ and $\mathsf C$}
        \UnaryInfC{$\mathsf{BA}a: \mathsf{CA}a:: \mathsf{AB}a: \mathsf{AC}a$}
    \BinaryInfC{$a:a:: \mathsf{AB}a: \mathsf{AC}a$}
        \AxiomC{$\mathsf A$ is an analogy}
        \UnaryInfC{$\mathsf{AB}a: \mathsf{AC}a:: \mathsf Ba: \mathsf Ca$}
    \RightLabel{t}
    \BinaryInfC{$a:a: \mathsf Ba: \mathsf Ca$}
    \UnaryInfC{$\mathsf B\equiv \mathsf C$.}
\end{prooftree} For the other direction, we compute:
\begin{prooftree}
    \AxiomC{$\mathsf B\equiv \mathsf C$}
    \UnaryInfC{$a:a:: \mathsf Ba: \mathsf Ca$}
    \RightLabel{$\mathsf A$ is an analogy}
    \UnaryInfC{$a:a:: \mathsf{AB}a: \mathsf{AC}a$}
        \AxiomC{$\mathsf A$ commutes with $\mathsf B$ and $\mathsf C$}
        \UnaryInfC{$\mathsf{AB}a: \mathsf{AC}a:: \mathsf{BA}a: \mathsf{CA}a$}
    \RightLabel{t}
    \BinaryInfC{$a:a:: \mathsf{BA}a: \mathsf{CA}a$}
    \UnaryInfC{$\mathsf{BA}a\equiv \mathsf{CA}a$.}
\end{prooftree}
\end{proof}

\begin{theorem}\label{t:F_G}
    \AxiomC{$\mathsf F\equiv \mathsf G$}
    \RightLabel{et.}
    \UnaryInfC{$\mathsf F:: \mathsf G$}
    \DisplayProof
\end{theorem}
\begin{proof}\hfill
\begin{prooftree}
    \AxiomC{$\mathsf F\equiv \mathsf G$}
    \RightLabel{s}
    \UnaryInfC{$\mathsf Fa: \mathsf Ga::a:a$}
       \AxiomC{inner reflexivity \prettyref{eq:e}}
       \RightLabel{e}
       \UnaryInfC{$a:a::b:b$}
           \AxiomC{$\mathsf F\equiv \mathsf G$}
           \UnaryInfC{$b:b:: \mathsf Fb: \mathsf Gb$}
           \RightLabel{t}
       \BinaryInfC{$a:a:: \mathsf Fb: \mathsf Gb$}
    \RightLabel{t}
    \BinaryInfC{$\mathsf Fa: \mathsf Fb:: \mathsf Ga: \mathsf Gb$}
    \UnaryInfC{$\mathsf F:: \mathsf G$.}
\end{prooftree}
\end{proof}

\subsection{Join equivalence}\label{§:Join}

Given $\mathsf F, \mathsf G:\mathfrak P\to \mathfrak P$, define their \textit{\textbf{(proportional) join equivalence}} by
\begin{align*} 
    \mathsf F\lor \mathsf G \quad:\Longleftrightarrow\quad a: \mathsf Fa::a: \mathsf Ga,\quad\text{for all $a\in P$}.
\end{align*} This can be depicted as follows:
\begin{center}
\begin{tikzpicture} 
    \node (a) {$a$};
    \node (a') [above=of a] {};
    \node (Fa) [left=of a'] {$\mathsf Fa$};
    \node (Ga) [right=of a'] {$\mathsf Ga$};
    \draw (a) to (Fa);
    \draw (a) to (Ga);
\end{tikzpicture}
\end{center}

\begin{fact}\label{f:lor_equiv} $\mathsf F\lor \mathsf G$ iff $\mathsf F\equiv \mathsf G$ in c-proportoids.
\end{fact}

\begin{corollary}\label{c:join_cong_pct} Join equivalence is a congruence for analogies on any ct-proportoid.
\end{corollary}
\begin{proof} A direct consequence of \prettyref{t:equiv_cong} and \prettyref{f:lor_equiv}.
\end{proof}

\prettyref{c:join_cong_pct} has rather strong assumptions as central permutation often fails in practice and it holds only for analogies on any ct-proportoid. The next result states that for join equivalence to be an equivalence relation (not a congruence), only transitivity is required:

\begin{fact} Join equivalence is an equivalence relation on any t-proportoid.
\end{fact}

The following result gives us a simple method to show that two mappings are \textit{not} join equivalent in case determinism \prettyref{eq:d} holds:\footnote{This is analogous to \prettyref{t:equiv_fix}.}

\begin{proposition}\label{p:join_fix} Join equivalent mappings have the same fixed points in d-proportoids.
\end{proposition}
\begin{proof} Given join equivalent mappings $\mathsf F, \mathsf G$, determinism implies that for any $a\in P$, in case $a$ is a fixed point of $\mathsf F$ thus satisfying $\mathsf Fa=a$,
\begin{align*} 
    a: \mathsf Fa::a: \mathsf Ga \quad\Longleftrightarrow\quad \mathsf Ga=a,
\end{align*} which means that $a$ is a fixed point of $\mathsf G$ as well. An analogous argument shows that every fixed point of $\mathsf G$ is a fixed point of $\mathsf F$.
\end{proof}

\begin{corollary}\label{c:F_lor_I} $\mathsf F\lor \mathsf I$ iff $\mathsf F = \mathsf I$ in d-proportoids.
\end{corollary}
\begin{proof} A direct consequence of \prettyref{p:join_fix}.
\end{proof}

The next observation shows that the converse of \prettyref{p:join_fix} fails in general since $S^k$and $S^\ell$ have the same fixed points (none) for all $k,\ell\geq 1$:

\begin{example}\label{e:S^k_S^ell_lor} In $(\mathbb N, ::)$ defined as in \prettyref{e:N}, we have $S^k\lor S^\ell$ iff $k=\ell$.
\end{example}

\subsection{Triangular relation}

Given $\mathsf F, \mathsf G:\mathfrak P\to \mathfrak P$, define the (\textit{\textbf{proportional}}) \textit{\textbf{triangular relation}} by
\begin{align*} 
    \mathsf F\;\triangledown\; \mathsf G \quad:\Longleftrightarrow\quad \mathsf F\lor \mathsf G \quad\text{and}\quad a: \mathsf Fa:: \mathsf Fa: \mathsf Ga \quad\text{and}\quad a: \mathsf Ga:: \mathsf Ga: \mathsf Fa,\quad\text{for all $a\in P$}.
\end{align*}

This can be depicted as follows (see the similarity to join equivalence in \prettyref{§:Join}):
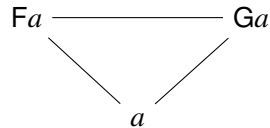
\begin{center}
\begin{tikzpicture} 
    \node (a) {$a$};
    \node (a') [above=of a] {};
    \node (Fa) [left=of a'] {$\mathsf Fa$};
    \node (Ga) [right=of a'] {$\mathsf Ga$};
    \draw (a) to (Fa);
    \draw (a) to (Ga);
    \draw (Fa) to (Ga);
\end{tikzpicture}
\end{center}

The following observation shows that the triangular relation is in general \textit{not} reflexive:

\begin{proposition}\label{p:F_triangledown_F} $\mathsf F\;\triangledown\; \mathsf F$ iff $\mathsf F=\mathsf I$ in d-proportoids.
\end{proposition}
\begin{proof} We have $\mathsf F\;\triangledown\; \mathsf F$ only if $a: \mathsf Fa:: \mathsf Fa: \mathsf Fa$, for all $a\in P$. By determinism, this is equivalent to $\mathsf Fa=a$, that is, $a$ is a fixed point of $\mathsf F$ for \textit{each} $a\in P$, which is equivalent to $\mathsf F= \mathsf I$. The other direction holds trivially.
\end{proof}

\begin{proposition}\label{p:triangledown->equiv}
    \AxiomC{$\mathsf F\;\triangledown\; \mathsf G$}
    \RightLabel{c.}
    \UnaryInfC{$\mathsf F\equiv \mathsf G$}
    \DisplayProof
\end{proposition}
\begin{proof} $\mathsf F\;\triangledown\; \mathsf G$ implies $\mathsf F\lor \mathsf G$ which implies $\mathsf F\equiv \mathsf G$ in all c-proportoids by \prettyref{f:lor_equiv}.
\end{proof}

\begin{example} As a direct consequence of \prettyref{e:S^k_S^ell_lor}, in $(\mathbb N, ::)$ we have $S^k\;\triangledown\; S^\ell$ iff $k=\ell$.
\end{example}

\subsection{Bowtie relation}\label{§:Bowtie_relation}

Given mappings $\mathsf F, \mathsf G:\mathfrak P\to \mathfrak P$, define the (\textit{\textbf{proportional}}) \textit{\textbf{bowtie relation}} by
\begin{align*} 
    \mathsf F\bowtie \mathsf G \quad:\Longleftrightarrow\quad a: \mathsf G\mathsf Fa::a: \mathsf{FG}a,\quad\text{for all $a\in P$},
\end{align*} which can be depicted as follows:
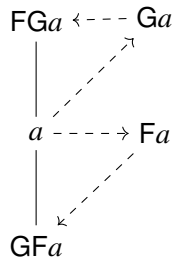
\begin{center}
\begin{tikzpicture} 
    \node (a) {$a$};
    \node (Fa) [right=of a] {$\mathsf Fa$};
    \node (GFa) [below=of a] {$\mathsf{GF}a$};
    \node (Ga) [above=of Fa] {$\mathsf Ga$};
    \node (FGa) [above=of a] {$\mathsf{FG}a$};
    \draw[->,dashed] (a) to (Fa);
    \draw[->,dashed] (Fa) to (GFa);
    \draw[-] (a) to (GFa);
    \draw[->,dashed] (a) to (Ga);
    \draw[->,dashed] (Ga) to (FGa);
    \draw[-] (a) to (FGa);
\end{tikzpicture}
\end{center} 

Notice that the bowtie relation is connected to the commutation of $\mathsf F$ and $\mathsf G$ and the next result shows that bowtie equivalent mappings commute with respect to proportional equivalence in c-proportoids satisfying central permutation:

\begin{proposition}\label{p:GF_equiv_FG} $\mathsf F\bowtie \mathsf G$ iff $\mathsf{GF}\equiv \mathsf{FG}$ in c-proportoids.
\end{proposition}
\begin{proof} The direction from right to left holds trivially. For the other direction, we have
\begin{prooftree} 
    \AxiomC{$\mathsf F\bowtie \mathsf G$}
    \UnaryInfC{$a: \mathsf G\mathsf Fa::a: \mathsf{FG}a$}
    \RightLabel{c}
    \UnaryInfC{$a:a: \mathsf G\mathsf Fa: \mathsf{FG}a$}
    \UnaryInfC{$\mathsf{GF}\equiv \mathsf{FG}$.}
\end{prooftree}
\end{proof}

\begin{fact} $\mathsf F\bowtie \mathsf I$ holds for every mapping $\mathsf F$.
\end{fact}
\begin{proof} An immediate consequence of reflexivity \prettyref{eq:r}.
\end{proof}

\begin{fact} The bowtie relation is reflexive and symmetric.
\end{fact}
\begin{proof} Follows from the reflexivity \prettyref{eq:r} and symmetry \prettyref{eq:s} of the analogical proportion relation.
\end{proof}

% \begin{problem}[] [Under which conditions] Is the bowtie relation transitive?
% \end{problem}

\begin{fact} For any mapping $\mathsf F$, we have $\mathsf F^k\bowtie \mathsf F^\ell$ for all $k,\ell\geq 0$. In particular, we have $\mathsf F\bowtie \mathsf F^2$ which means that $\mathsf F$ is idempotent with respect to bowtie equivalence.
\end{fact}
\begin{proof} A direct consequence of reflexivity \prettyref{eq:r}.
\end{proof}

\begin{example} In $(\mathbb N, ::)$, we have $S^k\bowtie S^\ell$ for all $k,\ell\geq 0$.
\end{example}

\subsection{Square equivalence}\label{§:Square_equivalence}

Given $\mathsf F, \mathsf G:\mathfrak P\to \mathfrak P$, define their (\textit{\textbf{proportional}}) \textit{\textbf{square equivalence}} by
\begin{align*} 
    \mathsf F\;\square\; \mathsf G \quad:\Longleftrightarrow\quad a: \mathsf Fa::b: \mathsf Gb \quad\text{and}\quad a:b:: \mathsf Fa: \mathsf Gb,\quad\text{for all $a,b\in P$},
\end{align*} which can be depicted as follows:
\begin{center}
\begin{tikzpicture} 
    \node (a) {$a$};
    \node (Fa) [right=of a] {$\mathsf Fa$};
    \node (Gb) [below=of Fa] {$\mathsf Gb$};
    \node (b) [below=of a,yshift=-0.05cm] {$b$};
    \draw (a) -- (b) -- (Gb) -- (Fa) -- (a);
\end{tikzpicture}
\end{center}

Applying $\mathsf F$ and $\mathsf G$ iteratively yields:
\begin{center}
\begin{tikzpicture} 
    \node (a) {$a$};
    \node (Fa) [right=of a] {$\mathsf Fa$};
    \node (b) [below=of a,yshift=-0.05cm] {$b$};
    \node (Gb) [below=of Fa] {$\mathsf Gb$};
    \node (FFa) [right=of Fa] {$\mathsf{FF}a$};
    \node (GGb) [below=of FFa] {$\mathsf{GG}b$};
    \node (FFFa) [right=of FFa] {\ldots};
    \node (GGGb) [right=of GGb] {\ldots};
    \draw (a) -- (b) -- (Gb) -- (Fa) -- (a);
    \draw (Fa) -- (FFa) -- (GGb) -- (Gb);
    \draw (FFa) -- (FFFa);
    \draw (GGb) -- (GGGb);
\end{tikzpicture}
\end{center}

Notice the similarity between square equivalence and the definition of an analogy in \prettyref{§:p-Analogy}, which immediately yields the following observation:

\begin{fact}\label{f:F_square_F} $\mathsf F\;\square\; \mathsf F$ iff $\mathsf F$ is an analogy, holds in any c-proportoid.
\end{fact}

\begin{fact} Proportional square equivalence is an equivalence relation for analogies on any t-proportoid.
\end{fact}
\begin{proof} Reflexivity follows from the fact that every mapping is an analogy by assumption (and see \prettyref{f:F_square_F}), symmetry follows from \prettyref{eq:s}, and transitivity follows from \prettyref{eq:t}.
\end{proof}

The next result shows that square equivalence is a very strong condition implying equivalence and join equivalence:

\begin{fact}\label{f:square->equiv}
    \AxiomC{$\mathsf F\;\square\; \mathsf G$}
    \RightLabel{.}
    \UnaryInfC{$\mathsf F\equiv \mathsf G$}
    \DisplayProof
\end{fact}

\begin{fact}\label{f:square->lor} 
    \AxiomC{$\mathsf F\;\square\; \mathsf G$}
    \RightLabel{.}
    \UnaryInfC{$\mathsf F\lor \mathsf G$}
    \DisplayProof
\end{fact}

\begin{example} As a direct consequence of \prettyref{e:S^k_S^ell_lor} and \prettyref{f:square->lor}, in $(\mathbb N, ::)$ we have $S^k\;\square\; S^\ell$ iff $k=\ell$.
\end{example}

\subsection{Complete square equivalence}

We define the \textit{\textbf{(proportional) complete square equivalence}} of $\mathsf F$ and $\mathsf G$ by
\begin{align*} 
    \mathsf F\boxtimes \mathsf G \quad:\Longleftrightarrow\quad a:b:: \mathsf Fa: \mathsf Gb \quad\text{and}\quad a: \mathsf Fa::b: \mathsf Gb \quad\text{and}\quad a: \mathsf Gb::b: \mathsf Fa\quad\text{for all $a,b\in P$}.
\end{align*} This can be depicted as follows:
\begin{center}
\begin{tikzpicture} 
    \node (a) {$a$};
    \node (Fa) [right=of a] {$\mathsf Fa$};
    \node (b) [below=of a] {$b$};
    \node (Gb) [right=of b,yshift=0.02cm] {$\mathsf Gb$};
    \draw (a) -- (b) -- (Gb) -- (Fa) -- (a);
    \draw (a) -- (Gb);
    \draw (b) -- (Fa);
\end{tikzpicture}
\end{center}

\begin{fact}\label{f:boxtimes->square} 
    \AxiomC{$\mathsf F\boxtimes \mathsf G$}
    \RightLabel{.}
    \UnaryInfC{$\mathsf F\;\square\; \mathsf G$}
    \DisplayProof
\end{fact}

\begin{fact} 
    \AxiomC{$\mathsf F\boxtimes \mathsf G$}
    \RightLabel{.}
    \UnaryInfC{$\mathsf F\equiv \mathsf G$}
    \DisplayProof
\end{fact}
\begin{proof} \hfill
\begin{prooftree}
    \AxiomC{$\mathsf F\boxtimes \mathsf G$}
    \RightLabel{\ref{f:boxtimes->square}}
    \UnaryInfC{$\mathsf F\;\square\; \mathsf G$}
    \RightLabel{\ref{f:square->equiv}}
    \UnaryInfC{$\mathsf F\equiv \mathsf G$.}
\end{prooftree}
\end{proof}

\section{Proportional circles}\label{§:Circles}

The following construction appears conceptually appealing:

\begin{definition} Given $a,b\in P$, we define the \textit{\textbf{(proportional) circle}} with center $a$ and ``radius'' $\overline{ab}$ by the solution set
\begin{align*} 
    \mathscr C_{ab} :=\mathscr S(a:b::a:x).
\end{align*}
\end{definition}

% FAIL
% \begin{fact} $\mathscr C_{ab}= \mathscr C_{ba}$.
% \end{fact}
% \begin{proof} An immediate consequence of p-symmetry.
% \end{proof}

The next result shows how to construct further solutions from a given one in any t-proportoid using circles:

\begin{theorem}\label{t:C_cd}
    \AxiomC{$d\in \mathscr S(a:b::c:x)$}
    \RightLabel{t.}
    \UnaryInfC{$\mathscr C_{cd}\subseteq \mathscr S(a:b::c:x)$}
    \DisplayProof
\end{theorem}
\begin{proof} We show the implication
\begin{align*} 
    e\in \mathscr C_{cd} \quad\Longrightarrow\quad e\in \mathscr S(a:b::c:x)
\end{align*} with the following derivation:
\begin{prooftree}
    \AxiomC{$d\in \mathscr S(a:b::c:x)$}
    \UnaryInfC{$a:b::c:d$}
        \AxiomC{$e\in \mathscr C_{cd}$}
        \UnaryInfC{$c:d::c:e$}
        \RightLabel{t}
    \BinaryInfC{$a:b::c:e$}
    \UnaryInfC{$e\in \mathscr S(a:b::c:x)$.}
\end{prooftree}
\end{proof}

% \prettyref{t:C_cd} can be visualized as follows:
% \missingfigure{}

\section{Function proportions}\label{§:Function_proportions}

Every relation on $P$ can be extended point-wise to a relation on functions on $P$ of same arity. For the analogical proportion relation extended to functions, we thus obtain the following definition:

\begin{definition}\label{d:EFGH} Given mappings $\mathsf{E,F,G,H}: \mathfrak P\to \mathfrak P$, we define the \textit{\textbf{function proportion relation}} by
\begin{align}\label{eq:EFGH} 
    \mathsf{E:F::G:H} \quad:\Longleftrightarrow\quad \mathsf Ea: \mathsf Fa:: \mathsf Ga: \mathsf Ha,\quad\text{for all $a\in P$.}
\end{align}
\end{definition}

\begin{fact}\label{f:EFGH} We have
\begin{align*} 
    &\mathsf{E:F::E:F} \quad\text{(reflexivity)},\\
    &\mathsf{E:F::G:H} \quad\Longleftrightarrow\quad \mathsf{G:H::E:F}\quad\text{(symmetry)},\\
    &\mathsf{E:F::G:H} \quad\Longleftrightarrow\quad \mathsf{F:E::H:G}\quad\text{(inner symmetry)}.
\end{align*} Similarly, all other properties of analogical proportions in \prettyref{d:proportoids} transfer to function proportions.
\end{fact}
% \begin{proof} Immediate consequences of \prettyref{d:proportoids}.
% \end{proof}

\begin{remark} \prettyref{f:EFGH} means that from any proportoid $\mathfrak P= (P, ::_{ \mathfrak P})$, we can construct the proportoid $$\mathfrak{P^P} := (P^P, ::_{ \mathfrak{P^P}})$$ of unary functions on $P$ with $::_{ \mathfrak{P^P}}$ defined point-wise as in \prettyref{eq:EFGH}.
\end{remark}

\begin{proposition}\label{p:EIFI} In any c-proportoid, we have
\begin{align*} 
    \mathsf{E:I::F:I} \quad\Longleftrightarrow\quad \mathsf E\equiv \mathsf F.
\end{align*}
\end{proposition}
\begin{proof} We have
\begin{align*} 
    \mathsf{E:I::F:I} 
        &\quad\Longleftrightarrow\quad \mathsf Ea:a:: \mathsf Fa:a,\quad\text{for all $a\in P$}\\
        &\quad\stackrel c\Longleftrightarrow\quad \mathsf Ea: \mathsf Fa::a:a,\quad\text{for all $a\in P$}\\
        &\quad\stackrel s\Longleftrightarrow\quad a:a:: \mathsf Ea: \mathsf Fa,\quad\text{for all $a\in P$}\\
        &\quad\Longleftrightarrow\quad \mathsf E\equiv \mathsf F.
\end{align*}
\end{proof}

The next result shows that proportional equivalence is compatible with function proportions:

\begin{proposition}\label{p:equiv->fp} 
    \AxiomC{$\mathsf E\equiv \mathsf F$}
        \AxiomC{$\mathsf G\equiv \mathsf H$}
        \RightLabel{t.}
    \BinaryInfC{$\mathsf{E:F::G:H}$}
    \DisplayProof
\end{proposition}
\begin{proof} \hfill
\begin{prooftree}
    \AxiomC{$\mathsf E\equiv \mathsf F$}
    \UnaryInfC{$a:a:: \mathsf Ea: \mathsf Fa$}
    \RightLabel{s}
    \UnaryInfC{$\mathsf Ea: \mathsf Fa::a:a$}
        \AxiomC{$\mathsf G\equiv \mathsf H$}
        \UnaryInfC{$a:a:: \mathsf Ga: \mathsf Ha$}
        \RightLabel{t}
    \BinaryInfC{$\mathsf{E:F::G:H}$.}
\end{prooftree}
\end{proof}

\begin{corollary}\label{c:square->fp}
    \AxiomC{$\mathsf E\;\square\; \mathsf F$}
        \AxiomC{$\mathsf G\;\square\; \mathsf H$}
        \RightLabel{t.}
    \BinaryInfC{$\mathsf{E:F::G:H}$}
    \DisplayProof
\end{corollary}
\begin{proof} \hfill
\begin{prooftree}
    \AxiomC{$\mathsf E\;\square\; \mathsf F$}
    \RightLabel{\ref{f:square->equiv}}
    \UnaryInfC{$\mathsf E\equiv \mathsf F$}
        \AxiomC{$\mathsf G\;\square\; \mathsf H$}
        \RightLabel{\ref{f:square->equiv}}
        \UnaryInfC{$\mathsf G\equiv \mathsf H$}
        \RightLabel{t, \prettyref{p:equiv->fp}}
    \BinaryInfC{$\mathsf{E:F::G:H}$.}
\end{prooftree}
\end{proof}

\begin{corollary}
    \AxiomC{$\mathsf E\boxtimes \mathsf F$}
        \AxiomC{$\mathsf G\boxtimes \mathsf H$}
        \RightLabel{t.}
    \BinaryInfC{$\mathsf{E:F::G:H}$}
    \DisplayProof
\end{corollary}
\begin{proof} \hfill
\begin{prooftree}
    \AxiomC{$\mathsf E\boxtimes \mathsf F$}
    \RightLabel{\ref{f:boxtimes->square}}
    \UnaryInfC{$\mathsf E\;\square\; \mathsf F$}
        \AxiomC{$\mathsf G\boxtimes \mathsf H$}
        \RightLabel{\ref{f:boxtimes->square}}
        \UnaryInfC{$\mathsf G\;\square\; \mathsf H$}
        \RightLabel{t, \prettyref{c:square->fp}}
    \BinaryInfC{$\mathsf{E:F::G:H}$.}
\end{prooftree}
\end{proof}

\begin{corollary}\label{c:E_triangledown_F} 
    \AxiomC{$\mathsf E\;\triangledown\; \mathsf F$}
        \AxiomC{$\mathsf G\;\triangledown\; \mathsf H$}
        \RightLabel{ct.}
    \BinaryInfC{$\mathsf{E:F::G:H}$}
    \DisplayProof
\end{corollary}
\begin{proof} \hfill
\begin{prooftree}
    \AxiomC{$\mathsf E\;\triangledown\; \mathsf F$}
    \RightLabel{c, \prettyref{p:triangledown->equiv}}
    \UnaryInfC{$\mathsf E\equiv \mathsf F$}
        \AxiomC{$\mathsf G\;\triangledown\; \mathsf H$}
        \RightLabel{c, \prettyref{p:triangledown->equiv}}
        \UnaryInfC{$\mathsf G\equiv \mathsf H$}
        \RightLabel{t, \prettyref{p:equiv->fp}}
    \BinaryInfC{$\mathsf{E:F::G:H}$.}
\end{prooftree}
\end{proof}

\begin{definition} We extend proportional identity from elements of $P$ to unary mappings on $P$ point-wise by
\begin{align*} 
    \mathsf F\stackrel{..}= \mathsf G \quad:\Longleftrightarrow\quad \mathsf Fa\stackrel{..}= \mathsf Ga,\quad\text{for all $a\in P$}.
\end{align*}
\end{definition}

The next result is a generalization of \prettyref{t:=_p->ap} from elements to functions:

\begin{proposition} 
    \AxiomC{$\mathsf E\stackrel{..}= \mathsf F$}
        \AxiomC{$\mathsf G\stackrel{..}= \mathsf H$}
        \RightLabel{t.}
    \BinaryInfC{$\mathsf{E:F::G:H}$}
    \DisplayProof
\end{proposition}
\begin{proof} \hfill
\begin{prooftree}
    \AxiomC{$\mathsf E\stackrel{..}= \mathsf F$}
    \UnaryInfC{$\mathsf Ea\stackrel{..}= \mathsf Fa$}
        \AxiomC{$\mathsf G\stackrel{..}= \mathsf H$}
        \UnaryInfC{$\mathsf Ga\stackrel{..}= \mathsf Ha$}
        \RightLabel{\ref{t:=_p->ap}}
    \BinaryInfC{$\mathsf Ea: \mathsf Fa:: \mathsf Ga: \mathsf Ha$}
    \UnaryInfC{$\mathsf{E:F::G:H}$.}
\end{prooftree}
\end{proof}

\section{Proportional similarity}\label{§:p-Similarity}

In this section, we shall introduce a notion of similarity in terms of analogical proportions:

\begin{definition} Given $a\in P$ and $b\in R$ and some set of functions $\Sigma:=\{\sigma_{ab}:P\to R\mid a,b\in P\}$, define
\begin{align*} 
    a\lesssim_\Sigma b \quad:\Longleftrightarrow\quad a:c::b:\sigma_{ab}c,\quad\text{for every $c\in P$},
\end{align*} and
\begin{align*} 
    a\approx_\Sigma b \quad:\Longleftrightarrow\quad a\lesssim_\Sigma b \quad\text{and}\quad b\lesssim_\Sigma a.
\end{align*} In case $a\approx_\Sigma b$, we say that $a$ and $b$ are \textit{\textbf{$\Sigma$-similar}}. 

% We extend similarity from elements to algebras by
% \begin{align*} 
%     \mathfrak P\lesssim_\Sigma\mathfrak R \quad:\Longleftrightarrow\quad \text{for each $a\in P$ there exists some $b\in R$ such that $a\approx_\Sigma b$,}
% \end{align*} and
% \begin{align*} 
%     \mathfrak P\approx_\Sigma\mathfrak R \quad:\Longleftrightarrow\quad \mathfrak P\lesssim_\Sigma\mathfrak R \quad\text{and}\quad \mathfrak R\lesssim_\Sigma\mathfrak P.
% \end{align*} 
\end{definition}

Notice that
\begin{align*} 
    a\lesssim_\Sigma b \quad&\Longleftrightarrow\quad \text{for every $c\in P$ there is some $d=\sigma_{ab}c\in R$ such that $a:c::b:d$}\\
        \quad&\Longrightarrow\quad \mathscr S(a:c::b:x)\neq\emptyset,\quad\text{for all $c\in P$}.
\end{align*}

\begin{proposition} $\Sigma$-similarity is reflexive and symmetric in any proportoid. Moreover, if
\begin{align}\label{eq:sigma} 
    \sigma_{ab}\sigma_{ca}d=\sigma_{cb}d
\end{align} holds for all $a,b,c,d\in P$, then $\Sigma$-similarity is an equivalence relation in any t-proportoid.
\end{proposition}
\begin{proof} Reflexivity follows from \prettyref{eq:r} which guarantees that for any $c\in P$ there is some $d:=c\in P$ such that $a:c::a:c$. Symmetry holds trivially. To prove transitivity, we proceed as follows. Suppose $a\approx_\Sigma b$ and $b\approx_\Sigma c$, which means that
\begin{align*} 
    &a:d::b:\sigma_{ab}d,\quad\text{for all $d\in P$},\\
    &b:e::c:\sigma_{bc}e,\quad\text{for all $e\in P$}.
\end{align*} Let $f\in P$ be an arbitrary element. We then have the following derivation:
\begin{prooftree}
    \AxiomC{$a:f::b:\sigma_{ab}f$}
        \AxiomC{$b:\sigma_{ab}f::c:\sigma_{bc}\sigma_{ab}f$}
        \RightLabel{t}
    \BinaryInfC{$a:f::c:\sigma_{bc}\sigma_{ab}f$}
    \RightLabel{\ref{eq:sigma}}
    \UnaryInfC{$a:f::c:\sigma_{cb}d$}
    \UnaryInfC{$a\approx_\Sigma c$.}
\end{prooftree}
\end{proof}

\begin{theorem}\label{t:approx} Let $\mathfrak P= (P, ::)$ be an it-proportoid. If there exists a set of functions $\Sigma$ satisfying \prettyref{eq:sigma} and the proportions
\begin{align}\label{eq:sigma_2} 
    \sigma_{ae}b:\sigma_{bf}a::\sigma_{cg}d:\sigma_{dh}c
\end{align} for all $a,b,c,d,e,f,g,h\in P$, then $\approx_\Sigma$ is a proportional congruence thus satisfying
\begin{prooftree}
    \AxiomC{$a\approx_\Sigma e$\quad $b\approx_\Sigma f$\quad $c\approx_\Sigma g$\quad $d\approx_\Sigma h$}
        \AxiomC{$a:b::c:d$}
        \RightLabel{.}
    \BinaryInfC{$e:f::g:h$}
\end{prooftree}
\end{theorem} 
\begin{proof} We have the derivations
\begin{prooftree}
    \AxiomC{$a\approx_\Sigma e$}
    \UnaryInfC{$a:b::e:\sigma_{ae}b$}
    \RightLabel{s}
    \UnaryInfC{$e:\sigma_{ae}b::a:b$}
        \AxiomC{$a:b::c:d$}
        \RightLabel{t}
    \BinaryInfC{$e:\sigma_{ae}b::c:d$}
        \AxiomC{$c\approx_\Sigma g$}
        \UnaryInfC{$c:d::g:\sigma_{cg}d$}
        \RightLabel{t}
    \BinaryInfC{$e:\sigma_{ae}b::g:\sigma_{cg}d$}
\end{prooftree} and
\begin{prooftree}
    \AxiomC{$b\approx_\Sigma f$}
    \UnaryInfC{$b:a::f:\sigma_{bf}a$}
    \RightLabel{y}
    \UnaryInfC{$a:b::\sigma_{bf}a:f$}
    \RightLabel{s}
    \UnaryInfC{$\sigma_{bf}a:f::a:b$}
        \AxiomC{$a:b::c:d$}
        \RightLabel{t}
    \BinaryInfC{$\sigma_{bf}a:f::c:d$}
        \AxiomC{$d\approx_\Sigma h$}
        \UnaryInfC{$d:c::h:\sigma_{dh}c$}
        \RightLabel{y}
        \UnaryInfC{$c:d::\sigma_{dh}c:h$}
        \RightLabel{t}
    \BinaryInfC{$\sigma_{bf}a:f::\sigma_{dh}c:h$.}
\end{prooftree} Now since we assume \prettyref{eq:sigma_2} and inner transitivity \prettyref{eq:i}, we have
\begin{prooftree}
    \AxiomC{$e:\sigma_{ae}b::g:\sigma_{cg}d$}
        \AxiomC{$\sigma_{ae}b:\sigma_{bf}a::\sigma_{cg}d:\sigma_{dh}c$}
        \RightLabel{i}
    \BinaryInfC{$e:\sigma_{bf}a::g:\sigma_{dh}c$}
        \AxiomC{$\sigma_{bf}a:f::\sigma_{dh}c:h$}
        \RightLabel{i}
    \BinaryInfC{$e:f::g:h$.}
\end{prooftree}
\end{proof}

% Es waere schoen, wenn ich die zusaetzliche Annahme in \prettyref{eq:sigma_2} im Beweis herleiten koennte, dh es waere schoen, wenn wir folgende Implikation haetten:
% \begin{prooftree}
%     \AxiomC{$a\approx_p e$\quad $b\approx_p f$\quad $c\approx_p g$\quad $d\approx_p h$}
%         \AxiomC{$a:b::c:d$}
%         \RightLabel{?}
%     \BinaryInfC{$\sigma_{ae}b:\sigma_{bf}a::\sigma_{cg}d:\sigma_{dh}c$}
% \end{prooftree}

\begin{fact} For any analogy $\mathsf A$ and any element $a$, we have $a\lesssim_\Sigma \mathsf Aa$.
\end{fact}

% The following result states that epimorphisms preserve $\Sigma$-similarity:

% \begin{proposition} For any epimorphism $\mathsf H:\mathfrak P\to \mathfrak R$,
% \begin{align*} 
%     a\approx_\Sigma b \quad\Longleftrightarrow\quad \mathsf Ha\approx_\Sigma \mathsf Hb,\quad\text{for all $a,b\in P$.}
% \end{align*}
% \end{proposition}
% \begin{proof} 
% \todo[inline]{}
% \end{proof}

% \begin{proposition} For any onto analogy $\mathsf A:\mathfrak P\to \mathfrak R$ on a ppt-triple $\mathfrak{PRP}$,
% \begin{align*} 
%     a\approx_\Sigma b \quad\Longleftrightarrow\quad \mathsf Aa\approx_\Sigma \mathsf Ab
% \end{align*} holds for all $a,b\in P$.
% \end{proposition}
% \begin{proof} Every analogy on a t-proportoid satisfies the strong proportion-preserving property by \prettyref{t:todo}.
% \todo[inline]{}
% \end{proof}

\section{Proportoids in universal algebra and predicate logic}\label{§:UA}

This paper is axiomatic in style in the sense that we do not study concrete realizations of the analogical proportion relation. However, in a series of papers the author has shown how a \textit{canonical} notion of an analogical proportion relation can be constructed from any algebra or structure in the sense of universal algebra \cite{Antic22,Antic23-22} and first-order logic \cite{Antic23-4}.

More formally, given a first-order language $L$ consisting of ranked function\footnote{Constant symbols are omitted and identified with 0-ary function symbols.} and relation symbols, an \textit{\textbf{$L$-structure}} consists of a non-empty set $P$ together with concrete functions and relations on $P$ corresponding to the function and relation symbols in $L$ (see e.g. \cite[§2]{Hinman05}). 

Given an $L$-structure $\mathfrak A= (A, \mathscr F, \mathscr R)$ with functions $\mathscr F$ and relations $\mathscr R$, we can define the \textit{\textbf{analogical proportion relation}} in $\mathfrak A$ --- in symbols,
\begin{align*} 
    a:b::_{ \mathfrak A} c:d
\end{align*} as in \cite{Antic23-4}. We do not want to go into technical details here. What is of interest here is that we can \textit{canonically} associate with any such $L$-structure $\mathfrak A= (A, \mathscr F, \mathscr R)$ a proportoid $\mathfrak A^{::} := (A, ::_{ \mathfrak A})$ by defining $::_{ \mathfrak A}$ as described in \cite{Antic23-4}. 

% In case we have two languages $L_{ \mathfrak A}$ and $L_{ \mathfrak B}$, an $L_{ \mathfrak A}$-structure $\mathfrak A= (A, \mathscr F, \mathscr R)$, and an $L_{ \mathfrak B}$-structure $\mathfrak B= (B, \mathscr G, \mathscr P)$, we can form the proportoids $\mathfrak A^{::}= (P, ::_{ \mathfrak A})$ and $\mathfrak B^{::}= (R, ::_{ \mathfrak B})$, which can then be connected by proportional homomorphisms $\mathfrak A^{::}\to \mathfrak B^{::}$. What is remarkable here is that with that procedure we obtain a \textbf{bilingual} framework of analogical proportions in the general setting of predicate logic.

Moreover, we can use algebras and structures to represent proportoids:

\begin{definition} Let $L$ be a set of function symbols. Then we call an $L$-algebra $\mathfrak A = (A,L^{ \mathfrak A})$ a \textit{\textbf{representation}} of a proportoid $(A, ::)$ iff there are 4-ary $L$-terms $s,t$ such that
\begin{align*} 
    a:b::c:d \quad\Longleftrightarrow\quad s^{ \mathfrak A}(a,b,c,d) = t^{ \mathfrak A}(a,b,c,d)
\end{align*} holds for all $a,b,c,d\in A$. A proportoid is \textit{\textbf{representable}} iff it has a representation.
\end{definition}

\begin{example} Every metric proportoid $M= ( \mathbb R, ::_\delta)$ arising from a metric space $(\mathbb R,\delta)$ is, by definition \prettyref{eq:ab_delta_cd}, representable.
\end{example}

\section{Summary}\label{§:Summary}

The results obtained so far reveal a non-trivial dependency between the axioms of the analogical proportion relation in \prettyref{d:proportoids} and the induced properties of proportoids. The purpose of this brief section is to make this dependency transparent by summarising which properties follow from which axioms:

\begin{table}[H]
\begin{tabularx}{\textwidth}{|c|X|}
    \hline
    \textbf{Axiom} & \textbf{Results}\\
    \hline
    % e
    inner reflexivity (e; \ref{eq:e}) & \prettyref{p:=_c}, \prettyref{p:=_p_=_c}, \prettyref{p:a=b->a=_pb}, \prettyref{t:=_p_congruence}, \prettyref{t:F_G}\\
    \hline
    % d
    determinism (d; \ref{eq:d}) & \prettyref{t:FIT}, \prettyref{t:SIT}, \prettyref{p:a=b->a=_pb}, \prettyref{t:equiv_fix}, \prettyref{c:F_equiv_I}, \prettyref{p:join_fix}, \prettyref{c:F_lor_I}, \prettyref{p:F_triangledown_F}\\
    \hline
    % c
    central permutation (c; \ref{eq:c}) & \prettyref{p:=_c}, \prettyref{p:e_f_g_h}, \prettyref{t:=_p_congruence}, \prettyref{t:Afa=_pfAa}, \prettyref{p:p-fixed}, \prettyref{f:diamond_fp}, \prettyref{f:diamond_con}, \prettyref{f:lor_equiv}, \prettyref{c:join_cong_pct}, \prettyref{p:triangledown->equiv}, \prettyref{p:GF_equiv_FG}, \prettyref{f:F_square_F}, \prettyref{p:EIFI}, \prettyref{c:E_triangledown_F}\\
    \hline
    % t
    transitivity (t; \ref{eq:t}) & Transitivity occurs in almost all results, which shows its importance in the theory of analogical proportions\\%\prettyref{t:A_sPPP}, \prettyref{t:SIT}, \prettyref{t:analogies_monoid}, \prettyref{f:sPPP}, \prettyref{p:=_c}, \prettyref{p:e_f_g_h}, \prettyref{p:ab_in_ker}, \prettyref{p:=_p_=_c}, \prettyref{p:a=b->a=_pb}, \prettyref{l:a_b_e_e}, \prettyref{t:=_p_congruence}, \prettyref{t:=_p->ap}, \prettyref{t:Afa=_pfAa}, \prettyref{l:FP}, \\
    \hline
    % i
    inner transitivity (i; \ref{eq:i}) & \prettyref{t:analogy}, \prettyref{p:diamond_equ_i}, \prettyref{p:diamond_equ}, \prettyref{t:equiv_cong}, \prettyref{t:approx}\\
    \hline
    % f
    functionality (f; \ref{eq:f}) & \prettyref{t:Afa=_pfAa}\\
    \hline
\end{tabularx}
\end{table}

\section{Conclusion}\label{§:Conclusion}

This paper introduced proportoids as sets endowed with a 4-ary analogical proportion relation satisfying a suitable set of axioms which are rooted in but different from Lepage's initial axiomatization. We then introduced proportional homomorphisms and their congruences and showed that they are related in the usual sense via kernels. Moreover, we introduced proportional analogies and showed that in transitive proportoids, they satisfy the strong proportion-preserving principle and are thus closely related to proportional homomorphisms. We showed how partial proportional analogies can be constructed from an enumeration and a selection function. We introduced a number of binary relations between unary mappings on proportoids. Finally, we introduced a notion of similarity in terms of analogical proportions.

From a mathematical point of view, it is interesting to further develop the mathematical theory of proportoids as initiated in this paper in analogy to other algebraic theories like, for example, lattice or category theory. In particular, proportoids can be combined with other relations or operations to form advanced proportoids; for example, we can combine a proportoid $(P, ::)$ with a semigroup $(P,\cdot)$ to form a ``semigroup proportoid'' $(P,\cdot, ::)$, given a suitable set of additional axioms which guarantee that $\cdot$ and $::$ interact properly.

From an artificial intelligence perspective, it is interesting to transfer the concepts introduced in this paper to settings relevant in AI-research. For example, in analogical logic programming \cite{Antic23-23} we study analogical proportions between logic programs and their structure-preserving mappings as introduced in this paper, which form a powerful symbiosis for logic program synthesis by analogy. In that context, proportional homomorphisms and analogies as introduced here correspond to logic program transformations preserving the proportional relationships between programs --- this can be interpreted as a form of learning novel logic programs by analogy, which is a promising novel approach to symbolic learning. Finally, conceptually comparing the notion of similarity introduced in \prettyref{§:p-Similarity} with the one in \cite{Antic23-2} appears interesting.

\section*{Acknowledgements}

I thank the referee for providing constructive criticism, which improved the quality of the paper.

\if\isdraft1\newpage\fi
\bibliographystyle{acm}
\bibliography{/Users/christianantic/Bibdesk/Bibliography,/Users/christianantic/Bibdesk/Publications_J,/Users/christianantic/Bibdesk/Publications_C,/Users/christianantic/Bibdesk/Preprints,/Users/christianantic/Bibdesk/Submitted}
\if\isdraft1\newpage

\section{Proportionally identical functions}

\begin{definition} Given mappings $\mathsf{F,G}: \mathfrak P\to \mathfrak R$, define their \textit{\textbf{proportional identity}} (or \textit{\textbf{p-identity}}) by
\begin{align*} 
    \mathsf F\stackrel{..}= \mathsf G \quad:\Longleftrightarrow\quad \mathsf Fa\stackrel{..}= \mathsf Ga,\quad\text{for all $a\in P$.}
\end{align*}
\end{definition}

\section{Commutative squares}

\begin{definition} We say that $a,b,c,d\in P$ form a \textit{\textbf{commutative square}} iff
\begin{align*} 
    a:b::c:d \quad\text{and}\quad a:c::b:d.
\end{align*}
\end{definition}
 
This can be depicted as follows:
\begin{center}
\begin{tikzpicture} 
    \node (a) {$a$};
    \node (c) [right=of a] {$c$};
    \node (d) [below=of c] {$d$};
    \node (b) [left=of d] {$b$};
    \draw (a) -- (b) -- (d) -- (c) -- (a);
\end{tikzpicture}
\end{center} Of course, in c-proportoids every analogical proportion is a commutative square. We will therefore mainly be interested in proportoids not satisfying central permutation \prettyref{eq:c}.

\section{Weak function proportions}

\begin{definition} Given mappings $\mathsf{E,F,G,H}:P\to P$, we define the \textit{\textbf{weak function proportion relation}} by
\begin{align*} 
    \mathsf{E:F\;w\;G:H} \quad:\Longleftrightarrow\quad \mathsf Ea: \mathsf Fb :: \mathsf Gc : \mathsf Hd,\quad\text{for some $a,b,c,d\in P$}.
\end{align*}
\end{definition}

\begin{fact} 
    \AxiomC{$\mathsf{E:F::G:H}$}
    \RightLabel{.}
    \UnaryInfC{$\mathsf{E:F\;w\;G:H}$}
    \DisplayProof
\end{fact}

\section{Proportional complete square equivalence}

\begin{definition} We define the \textit{\textbf{(proportional) complete square equivalence}} of $F$ and $G$ by
\begin{align*} 
    F\boxtimes G \quad:\Longleftrightarrow\quad a:b::Fa:Gb \quad\text{and}\quad a:Fa::b:Gb \quad\text{and}\quad a:Gb::b:Fa\quad\text{for all $a,b\in A$}.
\end{align*}
\end{definition}

Proportional complete square equivalence can be depicted as follows:
\begin{center}
\begin{tikzpicture} 
    \node (a) {$a$};
    \node (Fa) [right=of a] {$Fa$};
    \node (Gb) [below=of Fa] {$Gb$};
    \node (b) [left=of Gb] {$b$};
    \draw (a) -- (b) -- (Gb) -- (Fa) -- (a);
    \draw (a) -- (Gb);
    \draw (b) -- (Fa);
\end{tikzpicture}
\end{center}

\begin{fact} 
    \AxiomC{$F\boxtimes G$}
    \RightLabel{.}
    \UnaryInfC{$F\square G$}
    \DisplayProof
\end{fact}

\section{Proportional projection}

\begin{definition} We call any function $\pi : \mathfrak P^2\to \mathfrak R$ a \textit{\textbf{proportional projection}} iff it satisfies
\begin{align*} 
    a : b ::_{ \mathfrak{(P,R)}} \pi(a,b) : \pi(b,a), \quad\text{for all $a,b\in \mathbb P$}.
\end{align*}
\end{definition}

\begin{fact} The projection $\pi_1: \mathfrak P\to \mathfrak P$ is a proportional projection since
\begin{align*} 
    a : b :: \pi_1(a,b) : \pi_1(b,a) \quad\Longleftrightarrow\quad a:b::a:b,
\end{align*} which always holds by p-reflexivity. If the underlying proportoid is p-commutative, then $\pi_2$ is a proportional projection as well since
\begin{align*} 
    a : b :: \pi_2(a,b) : \pi_2(b,a) \quad\Longleftrightarrow\quad a:b::b:a.
\end{align*}
\end{fact}

\section{Proportional solution operator}

\begin{definition} A \textit{\textbf{(proportional) solution operator}} is any mapping $\mathsf S: \mathbb P^2\times \mathbb R\to \mathbb R$ satisfying
\begin{align*} 
    a : b ::_{ \mathfrak{(P,R)}} c : \mathsf S(a,b,c),\quad\text{for all $a,b\in \mathbb P$ and $c\in \mathbb R$}.
\end{align*}
\end{definition}

\begin{remark} A solution operator can exist only on proportoids where each proportional equation $a:b::c: \mathfrak x$ has a solution.
\end{remark}

\section{Proportional realization}

\begin{definition} Let $\mathfrak P$ and $\mathfrak R$ be proportoids, and let $a\in \mathbb P$ and $c\in \mathbb R$ be elements. We say that $c$ \textit{\textbf{(proportionally) realizes}} $a$ in $\mathfrak R$ --- written $a\, \mathsf R_{ \mathfrak{PR}}\, c$ --- iff for every $b\in \mathbb P$ there is some $d\in \mathbb R$ such that $a:b::_{ \mathfrak{PR}} c:d$.
\end{definition}

\section{Proportional ideals}

\begin{definition} Let $\mathfrak P$ be a proportoid. We call a subset $I\subseteq \mathbb P$ a \textit{\textbf{(proportional) ideal}} of $\mathfrak P$ --- in symbols, $I\lhd \mathfrak P$ --- iff it satisfies the following implication, for all elements in $\mathbb P$:
\begin{prooftree}
    \AxiomC{$a,b\in I$}
        \AxiomC{$a : b ::_{ \mathfrak P} a : c$}
    \RightLabel{.}
    \BinaryInfC{$c\in I$}
\end{prooftree}
\end{definition}

\begin{fact} Ideals are closed under circles, that is, if $a,b\in I$ then $\mathscr C_{ab}\subseteq I$.
\end{fact}

\begin{fact} Every singleton is trivially an ideal in any d-proportoid.
\end{fact}

It is reasonable to expect ideals and analogies to be related:

\begin{fact} For any analogy $\mathsf A$ and ideal $I$, we have
\begin{align*} 
    a,b\in I \quad\Longleftrightarrow\quad \mathsf Aa, \mathsf Ab\in I.
\end{align*}
\end{fact} 

\begin{remark} Notice that in a t-proportoid, every p-transitive consequence is contained in the p-ideal, that is, if $a:b::a:c$ and $a:c::a:d$ then $d\in I$.
\end{remark}

\begin{fact} A subset $I\subseteq A$ is a p-ideal iff for each p-circle $C_{a,b}$, with $a,b\in I$, we have $C_{a,b}\subseteq I$.
\end{fact} 

\section{Proportional projection}

\begin{definition} We call any function $\pi: \mathfrak P^2\to \mathfrak R$ a \textit{\textbf{proportional projection}} iff it satisfies
\begin{align*} 
    a : b ::_{ \mathfrak{(P,R)}} \pi(a,b) : \pi(b,a), \quad\text{for all $a,b\in \mathbb P$}.
\end{align*}
\end{definition}

\begin{fact} The projection $\pi_1: \mathfrak P\to \mathfrak P$ is a proportional projection since
\begin{align*} 
    a:b::\pi_1(a,b):\pi_1(b,a) \quad\Longleftrightarrow\quad a:b::a:b,
\end{align*} which always holds by p-reflexivity. If the underlying proportoid is p-commutative, then $\pi_2$ is a proportional projection as well since
\begin{align*} 
    a:b::\pi_2(a,b):\pi_2(b,a) \quad\Longleftrightarrow\quad a:b::b:a.
\end{align*}
\end{fact}

\section{Proportional copycat}

\begin{definition} A \textit{\textbf{(proportional) copycat}}\footnote{Homage to \cite{Hofstadter95a}.} is a pair of mappings $(C_{ \mathfrak P},C_{ \mathfrak R})$, where $C_{ \mathfrak P}: \mathbb P\times \mathbb R\to \mathbb P$ and $C_{ \mathfrak R}: \mathbb R\times \mathbb P\to \mathbb R$, satisfying
\begin{align*} 
    a : C_{ \mathfrak P}(a,b) ::_{ \mathfrak{(P,R)}} b: C_{ \mathfrak R}(b,a),\quad\text{for all $a\in \mathbb P$ and $b\in \mathbb R$}.
\end{align*} In case $\mathfrak P= \mathfrak R$ and $C_{ \mathfrak P}=C_{ \mathfrak R}$, we simply write $C_{ \mathfrak P}$ instead of $(C_{ \mathfrak P},C_{ \mathfrak P})$.
\end{definition}

\begin{example} The projection $\pi^2_1$ is a copycat on any e-proportoid by the assumed inner p-reflexivity.
\end{example}

% \subsection{Semicopycats}

\begin{definition} A \textit{\textbf{(proportional) semicopycat}} is a pair of mappings $(C_1,C_2)$, where $C_1: \mathbb P\to \mathbb P$ and $C_2: \mathbb R\times \mathbb P\to \mathbb R$, satisfying
\begin{align*} 
    a : C_1(a) ::_{ \mathfrak{(P,R)}} b : C_2(b,a),\quad\text{for all $a\in \mathbb P$ and $b\in \mathbb R$}.
\end{align*}
\end{definition}

% \subsection{Brainless copycats}

\begin{definition} A \textit{\textbf{brainless (proportional) copycat}} is a pair of mappings $(C_{ \mathfrak P},C_{ \mathfrak R})$, where $C_{ \mathfrak P}: \mathbb P\to \mathbb P$ and $C_{ \mathfrak R}: \mathbb R\to \mathbb R$, satisfying
\begin{align*} 
    a : C_{ \mathfrak P}(a) ::_{ \mathfrak{(P,R)}} b:C_{ \mathfrak R}(b),\quad\text{for all $a\in \mathbb P$ and $b\in \mathbb R$}.
\end{align*}
\end{definition}

\begin{fact} The brainless copycats on $(\mathbb N, ::)$ are given by $S^k$, for $k\geq 0$.
\end{fact}

\begin{fact} On c-proportoids, brainless copycats coincide with analogies.
\end{fact}

\section{Proportional proportoids}

\begin{definition} Given proportoids $\mathfrak P= (\mathbb P, ::)$ and $\mathfrak R= (\mathbb R, ::)$, we define
\begin{align*} 
    \mathfrak P \righttherefore \mathfrak R \quad:\Longleftrightarrow\quad \text{for all $a,b\in \mathbb P$ there are some $c,d\in \mathbb R$ such that $a:b::_{ \mathfrak{PR}} c:d$,}
\end{align*} and
\begin{align*} 
    \mathfrak P:: \mathfrak R \quad:\Longleftrightarrow\quad \mathfrak P \righttherefore \mathfrak R \quad\text{and}\quad \mathfrak R \righttherefore \mathfrak P
\end{align*} In case $\mathfrak P:: \mathfrak R$, we say that $\mathfrak P$ and $\mathfrak R$ are \textit{\textbf{proportional}}.
\end{definition}

\section{Proportional equivalence}

\begin{definition} Let $\mathfrak A = (\mathbb A,L^{ \mathfrak A})$, $\mathfrak B = (\mathbb B, L^{ \mathfrak B})$, $\mathfrak C = (\mathbb A, K^{ \mathfrak C})$, and $\mathfrak D = (\mathbb B, K^{ \mathfrak D})$ be structures with possibly different languages $L$ and $K$. We define
\begin{align*} 
    \mathfrak{(A,B)}\leqq_p \mathfrak{(C,D)} \quad:\Longleftrightarrow\quad \mathfrak{(A,B)} \models a:b::c:d \quad\text{implies}\quad \mathfrak{(C,D)} \models a:b::c:d,
\end{align*} for all $a,b\in \mathbb A$ and $c,d\in \mathbb B$, and
\begin{align*} 
    \mathfrak{(A,B)}\equiv_p \mathfrak{(C,D)} \quad:\Longleftrightarrow\quad \mathfrak{(A,B)}\leqq_p \mathfrak{(C,D)} \quad\text{and}\quad \mathfrak{(C,D)}\leqq_p \mathfrak{(A,B)}.
\end{align*} In case $\mathfrak{(A,B)}\equiv_p \mathfrak{(C,D)}$, we say that $\mathfrak{(A,B)}$ and $\mathfrak{(C,D)}$ are \textit{\textbf{proportionally equivalent}}.
\end{definition}

\newpage
\section*{TODOs}

\todo[inline]{$ker\, \mathsf A$ is a p-congruence?}

\todo[inline]{ueberpruefe, wo ``proportoid'' durch ``pproportoid'' zu ersetzen ist}

\todo[inline]{Aendere die Notation von $a:b::_{ \mathfrak P} c:d$ zu $\mathfrak P \models a:b::c:d$}

\newpage\section*{TAGEBUCH}

\section[7 April '26]{}

Ich moechte langsam wieder reinkommen und das AMAI-Review einarbeiten, damit ich mit dem Paper abschließen kann. 

\section[9 April '26]{}

Es faellt mir schwer, das Paper zu ueberarbeiten, weil's gerade so gar nicht in meinen Flow passt. Ich habe intern APs de facto aufgegeben und durch LP-Hom ersetzt, weswegen die Motivation fuer das Paper entfaellt. Ich moechte das Paper nur noch veroeffentlichen und dann vergessen. Das ist natuerlich keine gute Voraussetzung. Aber ich darf nicht vergessen, dass nur sehr wenig zu tun ist und dass ich an einem oder zwei Tagen fertig sein kann. Ich darf's daher einfach nur nicht zu lange aufschieben.

\section[12 April '26]{}

Das hier ist ein schoenes Paper --- ich bin froh, dass ich's geschrieben habe und freue mich darauf, wenn es publiziert ist.

\section[12 April '26]{}

Die Tabelle, die Rev B verlangt, nervt mich, weil sie keinen echten Nutzen bringt --- jedes Resultat listet doch schon alle Eigenschaften auf, die notwendig sind, um es zu beweisen!

Wo soll die Tabelle auch hinkommen?! Die Rede ist von ``final table'', also am Ende. 

\section[13 April '26]{}

So, ich komme nur sehr langsam weiter. Es ist schon wieder 17:33 Uhr und ich habe fast nichts gemacht. Ich sitze schon seit ueber einer Woche an diesem Paper und kann mich einfach nicht motivieren, zuegig zu schreiben, obwohl's nun wirklich nicht schwierig ist --- ich habe andere Paper viel schneller ueberarbeitet!

In \prettyref{§:I} muss ich den Bezug zur AI etwas deutlicher machen, da es im Moment eine sehr trockene Auflistung des Inhalts ist. Das sollte schnell gehen, denn das habe ich fuer APs schon oefters gemacht.

\fi
\end{document}